\documentclass[preprint,12pt]{elsarticle}

\usepackage{amssymb}
\usepackage{amsmath}
\usepackage{caption}
\usepackage{subcaption}

\usepackage{lineno}
\usepackage{amsmath,amsfonts,amsthm,bm} 

\usepackage[hyphens]{url}
\usepackage{hyperref}
\hypersetup{pdfauthor={Haoran Su}}
\usepackage{times}  
\usepackage{helvet} 
\usepackage{courier}  
\usepackage{placeins}

\usepackage{graphicx} 
\usepackage{amsmath}
\usepackage{subcaption,enumitem}
\usepackage{algpseudocode}
\usepackage{xcolor}
\usepackage{multirow}

\newcommand{\hs}[1]{{\color{black}{#1}}}

\newtheorem{remark}{Remark}
\usepackage[ruled,noend,linesnumbered]{algorithm2e}
\usepackage{booktabs}
\usepackage{multirow}
\usepackage{appendix}
\usepackage{etoolbox}
\usepackage[ruled,noend]{algorithm2e}
\usepackage{booktabs}
\usepackage{multirow}
\usepackage{amssymb}
\BeforeBeginEnvironment{appendix}{\clearpage}

\usepackage{soul}
\usepackage{cases}

\usepackage{natbib}  
\biboptions{numbers,square,comma,sort,compress}
\usepackage{caption} 
\journal{Transportation Research Part C}

\begin{document}

\begin{frontmatter}


\title{EMVLight: a Multi-agent Reinforcement Learning Framework for an Emergency Vehicle Decentralized Routing and Traffic Signal Control System}


\author[1, s]{Haoran Su}
\author[2, s]{Yaofeng D. Zhong}
\author[1]{Joseph Y.J. Chow}
\author[2]{Biswadip Dey}
\author[3]{Li Jin\corref{cor1}}
\ead{li.jin@sjtu.edu.cn}
\cortext[cor1]{Corresponding author}
\fntext[s]{Equal contribution}
\address[1]{Tandon School of Engineering, New York University, USA}
\address[2]{Siemens Corporation, Technology, USA}
\address[3]{UM Joint Institute and Department of Automation, Shanghai Jiao Tong University, China}
\begin{abstract}
\hs{Emergency vehicles (EMVs) play a crucial role in responding to time-critical calls such as medical emergencies and fire outbreaks in urban areas. Existing methods for EMV dispatch typically optimize routes based on historical traffic-flow data and design traffic signal pre-emption accordingly; however, we still lack a systematic methodology to address the coupling between EMV routing and traffic signal control. In this paper, we propose EMVLight, a decentralized reinforcement learning (RL) framework for joint dynamic EMV routing and traffic signal pre-emption. We adopt the multi-agent advantage actor-critic method with policy sharing and spatial discounted factor. This framework addresses the coupling between EMV navigation and traffic signal control via an innovative design of multi-class RL agents and a novel pressure-based reward function.
The proposed methodology enables EMVLight to learn network-level cooperative traffic signal phasing strategies that not only reduce EMV travel time but also shortens the travel time of non-EMVs.
Simulation-based experiments indicate that EMVLight enables up to a $42.6\%$ reduction in EMV travel time as well as an $23.5\%$ shorter average travel time compared with existing approaches.}
\end{abstract}

\end{frontmatter}


\section{Introduction}
Emergency vehicles (EMVs) include ambulances, fire trucks, and police cars, which respond to critical events such as medical emergencies, fire disasters, and public security crisis. Emergency response time is the key indicator of a city's incidents management ability and resiliency. Reducing response time saves lives and prevents property losses. For instance, \citet{berdowski2010global} indicates that the survivor rate from a sudden cardiac arrest without treatment drops 7\% - 10\% for every minute elapsed, and there is barely any chance to survive after 8 minutes. EMV travel time, the time interval for an EMV to travel from a rescue station to an incident site, accounts for a major portion of the emergency response time. However, overpopulation and urbanization have exacerbated road congestion, making it more challenging to reduce the average EMV travel time. Records \cite{end-to-end-response-times} have shown that even with a decline in average emergency response time, the average EMV travel time increased from 7.2 minutes in 2015 to 10.1 minutes in 2021 in New York City, an approximately 40\% increase over six years even accounting for post-Covid traffic conditions. Therefore, there is a severe urgency and significant benefit for shortening the average EMV travel time on increasingly crowded roads.

Existing work has studied strategies to reduce the travel time of EMVs by route optimization and traffic signal pre-emption \cite{Lu2019Literature, humagain2020systematic}. Route optimization refers to the search for a time-based shortest path. The traffic network (e.g., city road map) is modeled as a graph with intersections as nodes and road segments between intersections as edges. Based on the time a vehicle needs to travel through each edge (road segment), route optimization calculates an optimal route with the minimal EMV travel time \cite{humagain2020systematic}. In addition, as the EMV needs to be as fast as possible, the law in most places requires non-EMVs to yield to emergency vehicles sounding sirens, regardless of the traffic signals at intersections \cite{de1991lights}. Even though this practice gives the right-of-way to EMVs, it poses safety risks for vehicles and pedestrians at intersections \cite{grant2017human}. To address this safety concern, existing methods \cite{nelson2000impact, qin2012control, humagain2020systematic, huang2015design} have also studied traffic signal pre-emption which refers to the process of deliberately altering the signal phases at each intersection to prioritize EMV passage.
%
%

However, a major challenge for adaptive EMV operation is the coupling between
EMV route optimization and traffic light pre-emption \cite{humagain2020systematic}. As the traffic condition constantly changes, static route optimization can potentially become suboptimal as an EMV travels through the network; i.e. the traffic is highly time-dependent and exhibits transient properties during a dispatch \cite{Coogan2015Compartmental}. Moreover, traffic signal pre-emption has a significant impact on the traffic flow, which would change the fastest route as well. Thus, the optimal route should be updated with real-time traffic flow information, i.e., the route optimization should be solved in a dynamic (time-dependent) way. As an optimal route can change as an EMV travels through the traffic network, the traffic signal pre-emption would need to adapt accordingly. In other words, the subproblems of dynamic route optimization and traffic signal pre-emption are coupled and should be solved ideally simultaneously in real-time. Existing approaches have limited consideration to this coupling.

%
In addition, most of the existing models on emergency vehicle service have a single objective of reducing the EMV travel time \cite{Haghani2004Simulation, haghani2003optimization, panahi2008gis, shaaban2019strategy}. As a result, their traffic signal control strategies have an undesirable effect of increasing the travel time of non-EMVs. \hs{Non-EMVs on the path of approaching EMVs would pull over or stop, and they usually do not receive clear guidance from traffic signals on resuming their trips \cite{hsiao2018preventing}, causing unnecessary delay. Non-EMVs elsewhere are also likely indirectly and negatively impacted if adjacent intersections lack coordination to address the incurred delay \cite{humagain2020systematic}. Therefore, traffic signal control strategies accommodating both EMVs and non-EMVs need to be recommended.}

In this paper, we aim to perform route optimization and traffic signal pre-emption to not only reduce EMV travel time but also to reduce the average travel time of non-EMVs. In particular, we address the following two key challenges:
\begin{enumerate}
    \item \textbf{How to dynamically route an EMV to a destination under time-dependent traffic conditions in a computationally efficient way?} As the congestion level of each road segment changes over time, the routing algorithm should be able to update the remaining route as the EMV passes each intersection. Running the shortest-path algorithm each time the EMV passes through an intersection is not efficient. A computationally efficient decentralized routing algorithm is desired. \label{challenge_1}
    \item \textbf{How to coordinate traffic signals to not only reduce EMV travel time but reduce the average travel time of non-EMVs as well?} To reduce EMV travel time, only the traffic signals along the route of the EMV need to be altered. However, to further reduce average non-EMV travel time, traffic signals in the whole traffic network need to cooperate. \label{challenge_2}
\end{enumerate}

Reinforcement learning (RL), which gained significant traction in assorted domains of traffic signal control recently, has been extensively studied and proven effective for learning stochastic traffic conditions and dealing with randomness.
Thus, to tackle the above challenges, we propose \textbf{EMVLight}, a decentralized multi-agent reinforcement learning framework with a dynamic routing algorithm to control traffic signal phases for efficient EMV passage. Our experimental results demonstrate that EMVLight outperforms traditional traffic engineering methods and existing RL methods under two metrics - EMV travel time and the average travel time of all vehicles - on different traffic configurations. We extend the preliminary work \cite{Su2021EMVLight} by taking into account extra capacity of each road segments and the possibility of forming ``emergency lanes" for full speed EMV passage. In addition, we demonstrate EMVLight's performance on synthetic and real-world maps with extra capacities. We also present the difference in EMV routing between EMVLight and benchmark methods with an emphasize on the number of successfully formed emergency lanes.

Our contributions are threefold. First, we capture the emergency capacity in road segments for emergencies and incidents. We also propose a mathematical model to decide whether an emergency lane can be formed for full speed EMV passage based on emergency capacity and number of vehicles of a road segment.
Second, we incorporate a decentralized path-finding scheme for EMVs based on real time traffic information. Third, we propose to solve EMV routing and traffic signal control problems simultaneously in a multi-agent reinforcement learning framework. In particular, we set up different types of reinforcement learning agents based on the location of the EMV and design different rewards for each type. This leads to up to a $42.6\%$ reduction in EMV travel time as well as an $23.5\%$ shorter average travel time of all trips completed in the network compared with existing benchmark methods.

The rest of the paper is organized as follows. Section \ref{sec_related_work} reviews relevant literature. Section \ref{sec_preliminaries} introduces our definition of pressure and emergency capacity. Section \ref{sec_methodology} presents our EMVLight methodology, i.e.,  dynamic routing and reinforcement learning. Benchmark methods and experimental setup are presented in Section \ref{sec_experimentation}. Section \ref{sec_result} discussed and compared the performance of EMVLight and benchmark methods in terms of EMV travel time, average travel time of all vehicles as well as EMV route choices. We conclude in Section \ref{sec_conclusion} and share inspirations on future directions.

\section{Literature Review}\label{sec_related_work}

\textbf{Conventional routing optimization and traffic signal pre-emption for EMVs.}

%
%
Although, routing and pre-emption are coupled in reality, existing methods usually solve them separately. Many of the existing approached leverage Dijkstra's shortest path algorithm to get the optimal route \cite{wang2013development, Mu2018Route, cheng2016hybrid,cheng2016identification,cheng2016random,cheng2017body,you2019ct,you2019low,lyu2018super,you2022class,you2020unsupervised,you2021momentum,you2022simcvd,guha2020deep,you2022end,yang2020nuset,you2022incremental,you2022bootstrapping,liu2022gene,you2021megan,ma2022sparse,sun2022mirnf,ma2021good,ma2021undistillable, kwon2003route, JOTSHI20091}. An A* algorithm for ambulance routing was proposed by \citet{nordin2012finding}. However, this line of work assumed that the routes and traffic conditions are fixed and static and fails to address the dynamic nature of real-world traffic flows. Another line of work considered the change of traffic flows over time. \citet{ziliaskopoulos1993time} proposed a shortest-path algorithm for time-dependent traffic networks, but the travel time associated with each edge at each time step is assumed to be known prior. \citet{musolino2013travel} proposed different routing strategies for different times in a day (e.g., peak/non-peak hours) based on traffic history data at those times. However, in the problem of our consideration, routing and pre-emption strategies can significantly affect the travel time associated with each edge during the EMV passage, and the existing methods cannot deal with this kind of real-time changes. \citet{haghani2003optimization} formulated the dynamic shortest path problem as a mixed integer programming problem. \citet{koh2020real} have used RL for real-time vehicle navigation and routing. Related lines of work \cite{miller2000least, gao2006optimal, kim2005optimal, fan2005shortest, yang2014constraint, huang2012optimal, gao2012real, samaranayake2012tractable,cheng2016analysis,cheng2016banalysis,li2019novel,you2021sbilsan,chen2021self,nie2012optimal,you2018structurally,lyu2019super, thomas2007dynamic} tackled the optimal adaptive routing problem in a variety of stochastic and time-dependent settings. However, none of these studies have considered the traffic signal pre-emption coupling, nor did they took the context of EMV passage into account solving shortest path problems.

%
%
%
Once an optimal route for the EMV has been determined, traffic signal pre-emption is deployed to further reduce the EMV travel time. A common pre-emption strategy is to extend the green phases of green lights to let the EMV pass each intersection along a fixed optimal route \cite{wang2013development, bieker2019modelling}. Pre-emption strategies for multiple EMV requests were introduced by \cite{Asaduzzaman2017APriority}.
\citet{wu2020emergency} approached the emergency vehicle lane clearing from an microscopic motion planning perspective and \citet{hosseinzadeh2022mpc} investigated an EMV-centered traffic control scheme through multiple intersections to alleviate traffic congestion. These work scrutinized strategies on pre-emption, but they had not considered EMV's dynamic routing which leads to the optimal path.

%
%

Please refer to \citet{Lu2019Literature} and \citet{humagain2020systematic} for a thorough survey of conventional routing optimization and traffic signal pre-emption methods. We would also like to point out that the conventional methods prioritize EMV passage and have significant disturbances on the traffic flow which increases the average non-EMV travel time.
%
%
%
%

\textbf{RL-based traffic signal control.} 
Deep learning based techniques have improved the multi-domain study in recent years \cite{chen2021adaptive,you2020towards,you2021mrd,you2021self,liu2021auto,liu2021aligning,xu2021semantic,you2020contextualized,you2021knowledge}.
Traffic signal pre-emption only alters the traffic phases at the intersections where an EMV travels through. However, to reduce congestion, traffic phases at nearby intersections also need to be changed cooperatively. The coordination of traffic signals to mitigate traffic congestion is referred to as traffic signal control which has been addressed by leveraging deep RL in a growing body of work. 
\citet{abdulhai2003reinforcement} is among the first to explore Q-learning, which is a model-free reinforcement learning method to learn the optimal action in a particular state, for adaptive traffic signal control. The idea was outlined for an isolated traffic signal controller as well as a network of traffic signal controllers. However, the size of the state representation grows exponentially as more traffic signals are considered and the learning process becomes extremely expensive. Only the isolated scenario was numerically shown to have an advantage over traditional methods. 
To address this curse of dimensionality, \citet{prashanth2010reinforcement} introduced feature-based state representations, which grows linearly in the number of lanes considered. The improved state representation reduces computational complexity and improves performance. \citet{el2013multiagent} avoided the curse of dimensionality by introducing game-theoretic approaches into Q-learning where each agent converges to the best response policy to all neighbors' policies.

The increasing compute power in recent years enables the training of deep neural networks, in particular, the deep Q-learning RL method. Van der Pol et al. \citet{van2016coordinated} are the first to incorporate deep Q network (DQN) into traffic jam assessment and extend it to coordinated adjacent intersections. Based on that inspiration, a line of work proposes to leverage the Q-learning framework for traffic signal control.
CoLight~\cite{wei2019colight} facilitated communication and cooperation between nearby traffic signals by graph attentional networks.
FRAP~\cite{zheng2019frap} improved RL-based methods by adding a phase competition model.
PressLight~\cite{wei2019presslight} incorporated the method of max pressure, which aggressively selects the traffic signal phase with maximum pressure to smooth congestion\cite{varaiya2013max,LI2019Backpressure, LEVIN2020maxpressure, WANG2022Learning, Lazar2021Routing}, into the reward design and shows improvement over traditional reward design.
\citet{ThousandLights} combined the idea from FRAP~\cite{zheng2019frap} and PressLight~\cite{wei2019presslight} to achieve city-level traffic signal control.
\citet{Zang_Yao_Zheng_Xu_Xu_Li_2020} leveraged meta-learning algorithms to speed up Q-learning for traffic signal control.
This line of Q-learning methods assumed the traffic signal control agents to be homogeneous and the neural network parameters are shared among all the agents, which enables the methods to scale up. However, this assumption makes it hard to adapt these methods to non-regular traffic grids where different intersections might have different numbers of incoming and outgoing links. 

The actor-critic method is another category of reinforcement learning methods which does not suffer from the curse of dimensionality challenge in multi-agent settings.
\citet{aslani2017adaptive} studied actor-critic methods for adaptive traffic signal control under different traffic disruption scenarios and show that actor-critic controllers outperforms Q-learning-based controllers.
\citet{xu2021hierarchically} proposed a hierarchical actor-critic method to encourage cooperation between intersections.
\citet{chu2019multi} leveraged a multi-agent independent advantage actor-critic algorithm for adaptive traffic signal control. 
\citet{Ma2020Feudal} introduced a manager-worker hierarchy and split the entire traffic networks into unique regions, so that it can manage traffic conditions locally based on the actor-critic framework.
Existing RL-based traffic control methods focus on reducing the congestion in the traffic network and are not designed specifically for EMV pre-emption. In contrast, our RL framework is built upon state-of-the-art ideas such as max pressure and is designed to reduce both the EMV travel time and the overall congestion. 
The centralized training with decentralized execution framework empowers the scheme to be compatible with stochastic traffic settings and realize negligible communication and synchronization cost among vehicles. Under the actor-critic framework, \citet{Mo2020CVLight} investigated the traffic signal control strategy based on connected vehicle communications.
We recommend \cite{RLSurvey2022Mohammad,wei2019survey} for a comprehensive and systematic review on RL-based traffic signal control methods.
Although RL methods have been extensively studied on the particular problem of traffic signal control and proven beneficial and effective, no existing study has considered EMV passages as well as their negative impacts on non-EMVs traffic flow.

\textbf{Intra-link EMV traversal strategies.} Recent advances of vehicle-to-vehicle communication technology \cite{LeBrun2005Knowledge} have drawn attention to designing optimal link level traversal strategies for EMVs in congested roads. \citet{Agarwal2016V2V} proposed Fixed Lane Strategy (FLS) and Best Lane Strategy (BLS) for EMVs to travel through a congested urban roadway. Based on the findings of \cite{Agarwal2016V2V}, \citet{Insaf2019Emergency} studied the performance of both strategies from a communication perspective and conclude that FLS outperforms BLS when speed variance is small. \citet{Hannoun2019Facilitating, Hannoun2021Sequential} designed and improved a semi-automated warning system to instruct downstream traffic to yield using mixed integer linear programming. The proposed strategy adapts to varying market penetration rate of connected vehicles and is validated to be computationally efficient for the real-time deployment. \citet{su2021dynamic} suggested the dynamic queue-jump lane (DQJL) strategy using multi-agent reinforcement learning.

\section{Preliminaries}\label{sec_preliminaries}
In this section, we introduce relevant preliminary terms and definitions which facilitate the problem formulation.

\subsection{Traffic map, movement and signal phase}
\begin{figure}[h]
\includegraphics[width=0.65\linewidth]{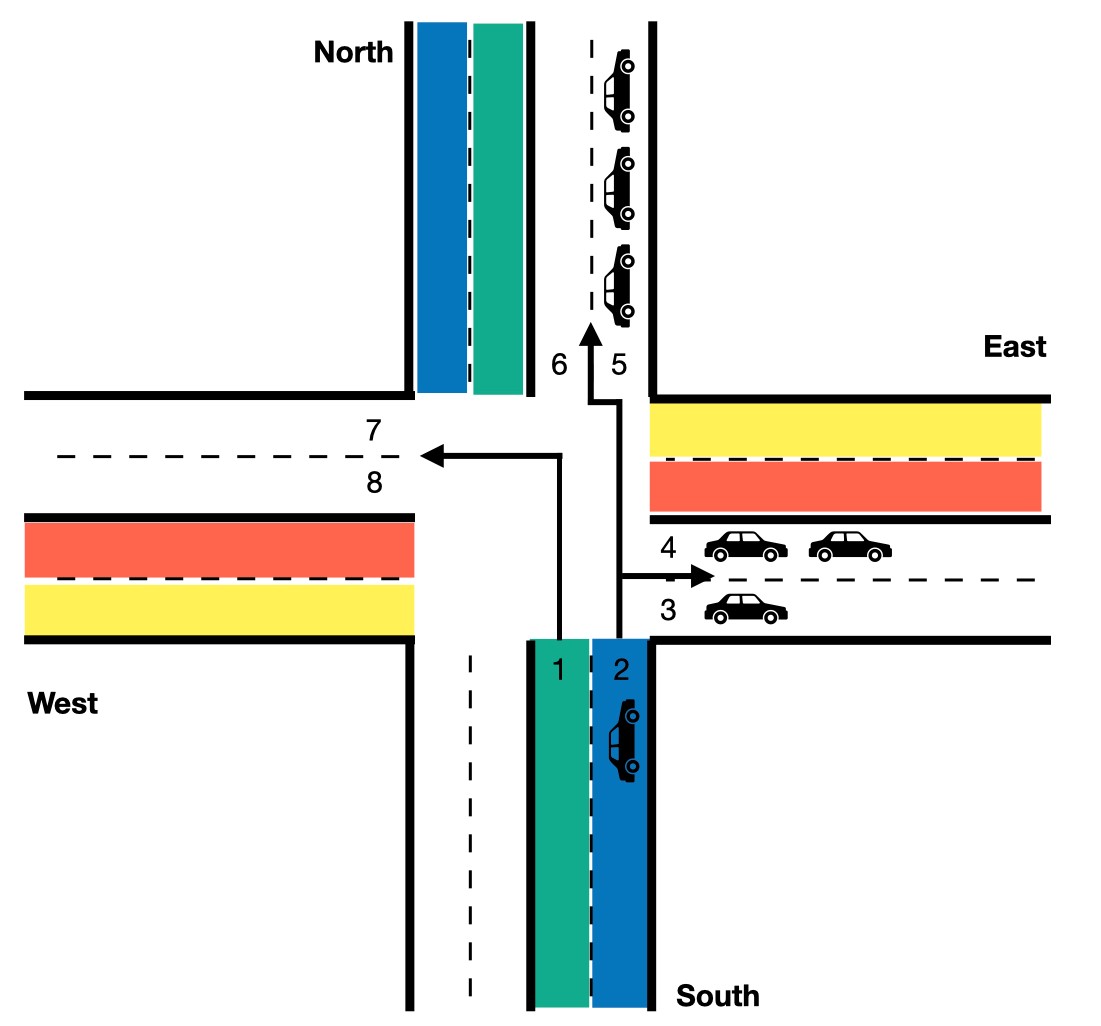}
\centering
\caption{Traffic movements of a four-way two-lane bidirectional intersection.}
\label{fig_movements}
\end{figure}
A traffic map can be represented by a graph $G(\mathcal{V}, \mathcal{E})$, with intersections as nodes and road segments between intersections as edges. We refer to a one-directional road segment between two intersections as a link. A link has a fixed number of lanes, denoted as $h(l)$ for lane $l$. Vehicles are allowed to switch lane between two intersections. Fig.~\ref{fig_movements} shows 8 links and each link has 2 lanes. 
    
A traffic movement $(l,m)$ is defined as the traffic traveling across an intersection from an incoming lane $l$ to an outgoing lane $m$. The intersection shown in Fig.~\ref{fig_movements} has 24 permissible traffic movements. As an example, vehicles on lane 1 are turning left, and vehicles on lane 2 may go straight or turn right. After turning into the link, vehicles will enter either lane. Thus, the incoming South link has the potential traffic movements of $\{(1, 7), (1, 8), (2, 5), (2, 6), (2, 3), (2, 4)\}$. The set of all permissible traffic movements of an intersection is denoted as $\mathcal{M}$.

\begin{figure}[t]
    \includegraphics[width=\linewidth]{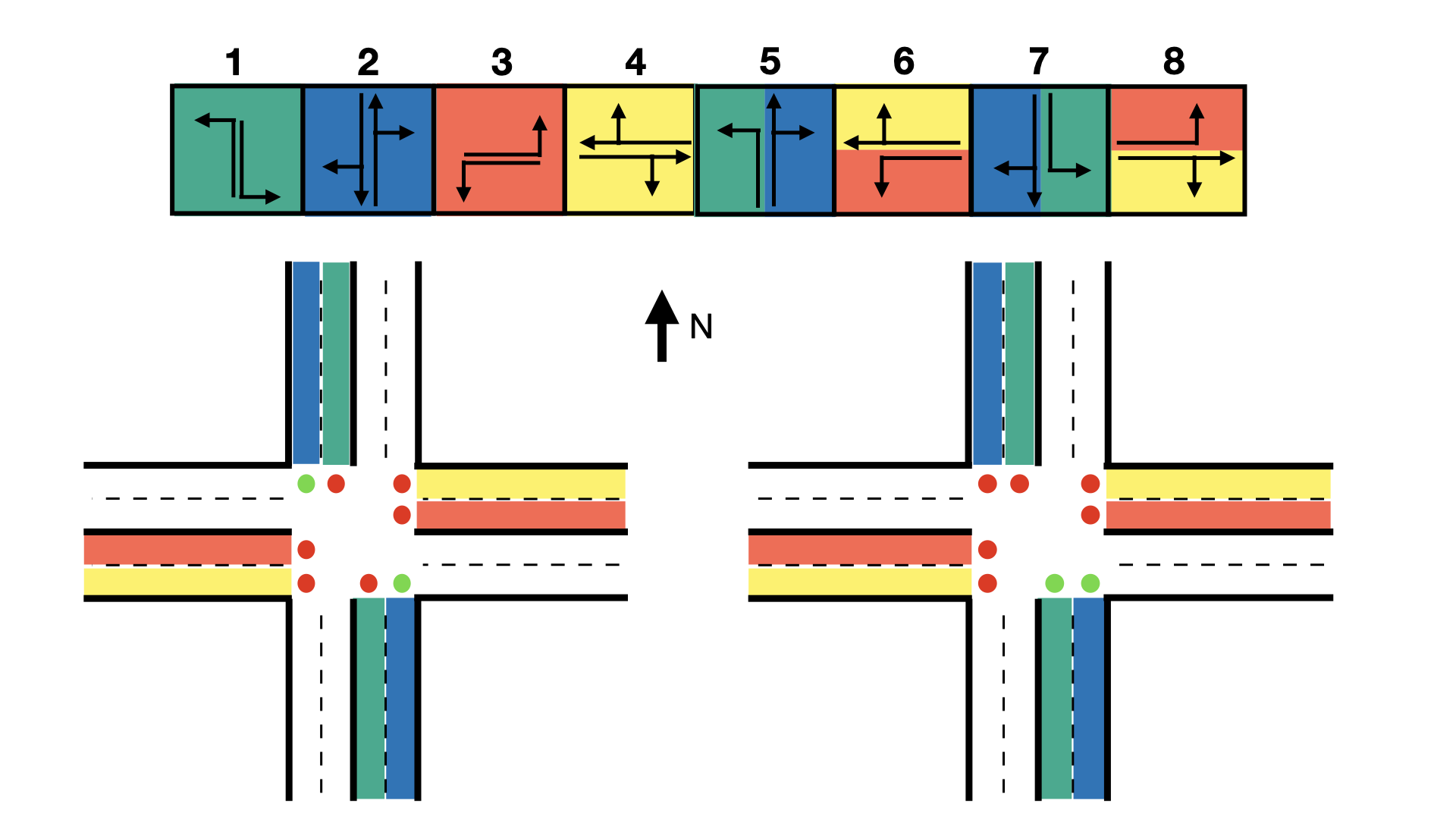}
    \centering
  \caption{\emph{Top}: 8 signal phases; \emph{Left}: phase \#2 illustration; \emph{Right}: phase \#5 illustration.
  }
  \label{fig_phases}
  \end{figure}

A traffic signal phase is defined as the set of permissible traffic movements. 
As shown in Fig.~\ref{fig_phases}, an intersection with 4 links has 8 phases.
%
%
\subsection{Pressure}
The pressure of an incoming lane $l$ measures the unevenness of vehicle density between lane $l$ and corresponding out going lanes in permissible traffic movements. The vehicle density of a lane is $x(l)/x_{max}(l)$, where $x(l)$ is the number of vehicles on lane $l$ and $x_{max}(l)$ is the vehicle capacity on lane $l$, which is related to the length of a lane. Then the pressure of an incoming lane $l$ is 
\begin{equation} 
    w(l) = \left|\frac{x(l)}{x_{max}(l)} - \sum_{\{m|(l, m)\in \mathcal{M}\}}\frac{1}{h(m)}\frac{x(m)}{x_{max}(m)}\right|,
    \label{eqn:lane_pressure}
\end{equation}
where $h(m)$ is the number of lanes of the outgoing link which contains $m$. In Fig.~\ref{fig_movements}, $h(m)=2$ for all the outgoing lanes. An example for Eqn.~\eqref{eqn:lane_pressure} is shown in Fig.~\ref{fig_movements}.

Taking the intersection's traffic conditions shown in Fig.\ref{fig_movements} as an example, assuming the maximum capacity for each lane is 5 vehicles, we can calculate the lane pressure for lane 2 to be
\begin{equation*}
    w(2) = \left| \underbrace{\frac{1}{5}}_{\textrm{lane 2}} - \frac{1}{2}(\underbrace{\frac{1}{5} + \frac{2}{5}}_{\textrm{lane 3 and 4}}) - \frac{1}{2}(\underbrace{\frac{3}{5} + \frac{0}{5}}_{\textrm{lane 5 and 6}})\right| = \frac{2}{5}
\end{equation*}


The pressure of an intersection indicates the unevenness of vehicle density between incoming and outgoing lanes in an intersection. Intuitively, reducing the pressure leads to more evenly distributed traffic, which indirectly reduce congestion and average travel time of vehicles. EMVLight defines pressure of an intersection in EMVLight as the average of the pressure of all incoming lanes,
\begin{equation*}\label{eq:EMVLight_pressure}
    P_{i} = \frac{1}{|\mathcal{I}_i|}\sum _{l\in \mathcal{I}_i} w(l),
\end{equation*}
where $\mathcal{I}_i$ represents the set of all incoming lanes of intersection $i$. According to such definition, the intersection pressure shown in Fig.\ref{fig_movements} is computed to be $\frac{25}{80}$.

PressLight \cite{wei2019presslight} assumes that traffic movements are lane-to-lane, i.e., vehicles in one lane can only move into a particular lane in a link. Because of the lane-to-lane assumption, in PressLight, the pressure is defined per movement. PressLight defines the pressure of a movement as the difference of the vehicle density between an incoming lane $l$ and the outgoing lane $m$, i.e., 
\begin{equation*}
    w^{*}(l, m) = \frac{x(l)}{x_{max}(l)} - \frac{x(m)}{x_{max}(m)},
\end{equation*}
For instance, lane 2, shown in Fig.\ref{fig_movements}, carries $w^{*}(2, 3)$, $w^{*}(2, 4)$, $w^{*}(2, 5)$, $w^{*}(2, 6)$, $w^{*}(2, 7)$ and $w^{*}(2, 8)$. Taking the permissible traffic movement from lane 2 to lane 4 as an example, we can get the pressure for this movement as
\begin{equation*}
    w^{*}(2, 4) = \frac{1}{5} - \frac{2}{5} = -\frac{1}{5}
\end{equation*}

PressLight then defines the pressure of an intersection $i$ as the absolute value of the sum of pressure of movements of intersection $i$, i.e., 
When calculating the pressure intersection, PressLight has
\begin{equation*}\label{eq:PressLight_pressure}
    P^{*}_{i} = \left|\sum _{(l, m)\in \mathcal{M}_i} w^{*}(l, m)\right|,
\end{equation*}
where $\mathcal{M}_i$ is the set of permissible traffic movements of intersection $i$. According to PressLight's definition of intersection pressure, Fig.\ref{fig_movements} shows an intersection with pressure of 6.

\paragraph{Pressure in our work}
EMVLight assumes a lane-to-link style traffic movement as vehicles can enter either lane on the target link, see Fig.\ref{fig_movements}. 
Following lane pressure defined in \eqref{eqn:lane_pressure},

\paragraph{Comparison}
The first difference between the two definitions is that $w^{*}(l, m)$ can be both positive or negative, but $w(l)$ can only take positive values that measures the unevenness of the vehicle density in the incoming lane and that of the corresponding outgoing lanes. We take the absolute value since the direction of pressure is irrelevant here, and the goal of each agent is to minimize this unevenness. The second difference is that at the intersection level, $P^{*}_{i}$ takes a sum but $P_{i}$ takes an average. The average is more suitable for our purpose since it scales the pressure down and the unit penalty for normal agents would be relatively large as compared to rewards for pre-emption agents (Eqn. (3)).  This design puts the efficient passage of EMV vehicles at the top priority. Our experimentation results, presented in Sec.\ref{sec_result} indicate the proposed pressure design produces a more robust reward signal during training and outperforms PressLight in congestion reduction.

\subsection{Emergency capacity}

\begin{figure}[ht]
\includegraphics[width=0.85\linewidth]{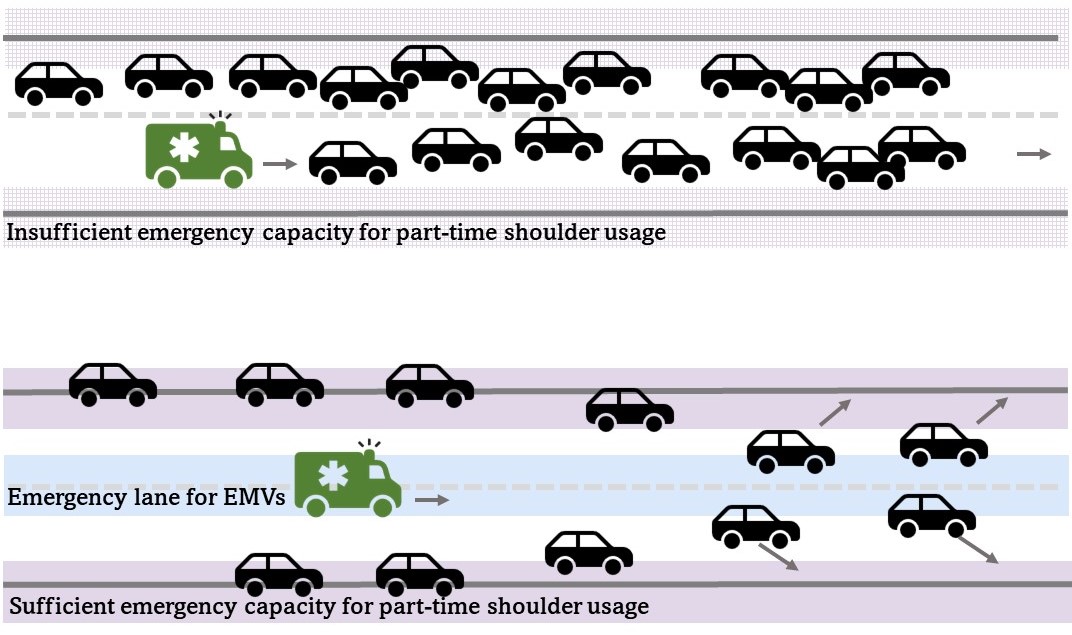}
\centering
\caption{A demonstration of an emergency lane for EMV passage. \emph{Top}: the EMV has to follow the congested queue due to insufficient emergency capacity;
\emph{Bottom}: the EMV is traversing on the yielded emergency lane due to sufficient emergency capacity.}
\label{fig_yielding}
\end{figure}

A roadway segment may have additional capacity, e.g. shoulder, parking lanes, bike lanes, dedicated to providing extra space that can be used under emergencies and incidents.
In the presence of EMVs, existing vehicles are allowed to pull-over or park on the shoulders temporarily, forming an \emph{emergency lane} for the emergency vehicle to pass. An emergency lane is an lane formed between original lanes dedicated for EMV passage, see Fig.\ref{fig_yielding} bottom. EMVs are assumed to travel freely on the emergency lane through the dense traffic. This is referred as an \emph{emergency yielding} and non-EMVs are experiencing an \emph{part-time shoulder use} \cite{part-time-shoulder-use}.

Adaptive traffic management strategies based on part-time shoulder use, such as dynamic hard shoulder running (D-HSR) \cite{Jiaqi2016Dynamic}, have proven beneficial and cost-effective for such scenarios. 

We use an intuitive mathematical model to determine whether such an emergency lane for EMV passage can be established. 
First, we define \emph{emergency capacity} $C_{i}^{\textrm{EC}}$ of a link $i$ to be the additional capacity in the link for emergencies and incidents. The emergency capacity depends on the segment's shoulder width, lane width, geometric clearance and other factors. We say that a link is \emph{emergency-capacitated} if it has a nonzero emergency capacity.
In order to express the maximum number of vehicles allowed in a link, for forming an emergency lane, we assume there are $n_i$ non-EMVs in the link $i$ with an average speed of $s_i$. We also assume that the normal capacity $k_i$ is evenly distributed among $l_i$ lanes so that each lane has a capacity of $k_i/l_i$. 
The overall capacity of the link is then $k_i + C_{i}^{\textrm{EC}}$. 
To form an emergency lane for EMV passage with maximum speed, all the non-EMVs need to move out of the emergency lane. Thus the maximum number of vehicles allowed is $k_i + C_{i}^{\textrm{EC}}- k_i/l_i$. As a result, the travel speed of the EMV is 
\begin{equation}\label{eqn:EMV_speed}
s_{i}^{\textrm{EMV}} = \begin{cases}
s_{f} & n_{i} \leq k_{i} + C_{i}^{\textrm{EC}} - \frac{k_i}{l_i},\\
s_{i} & \textrm{else,}
\end{cases}
\end{equation}
where $s_{f}$ represents the maximum speed allowed for EMVs.
%

\section{Methodology}\label{sec_methodology}
In this section, we elaborate the methodology of EMVLight. We begin with implementing a decentralized shortest path onto EMV navigation in traffic networks, and then incorporate it into the proposed multi-class RL agent design. Subsequently, we introduce the multi-agent advantage actor-critic framework as well as the RL training workflow in details.
\subsection{Decentralized Routing for EMVLight}
Dijkstra's algorithm is an algorithm that finds the shortest path between a given node and every other nodes in a graph. The time-based Dijkstra's algorithm finds the fastest path and has been widely used for EMV routing. In order to find such a path, the EMV travel time along each link need to be estimated first and we refer to it as the \emph{intra-link travel time.}
Dijkstra's algorithm takes as input the traffic graph, the intra-link travel time and a destination, and can return the time-based shortest path as well as estimated travel time from each intersection to the destination. The latter is usually referred to as the \emph{estimated time of arrival} (ETA) of each intersection.



In a traffic network, the intra-link travel time usually depends on the link's emergency capacity and number of vehicles in that link. In our model, this dependency is captured by EMV speed Eqn.~\eqref{eqn:EMV_speed}. The intra-link travel time is then calculated as the link length divided by the EMV speed. 
However, traffic conditions are constantly changing and so does EMV travel time along each link. Moreover, EMV pre-emption techniques alter traffic signal phases, which will significantly change the traffic condition as the EMV travels. The pre-determined shortest path might become congested due to stochasticity and pre-emption. Thus, updating the optimal route dynamically can facilitate EMV passage. One option is to run Dijkstra's algorithm repeatedly as the EMV travels through the network in order to take into account the updated EMV intra-link travel time.
\hs{However, this requires global traffic information of the entire traffic network throughout EMVs' trips. Even when a centralized controller is established to navigate the EMV, the synchronization and communication cost grows exponentially when the network size increases and the nonscability disallows the centralized scheme to be real-time for the navigation. 

Decentralized routing approaches \cite{ADACHER2014routeguidance, Chen2006Riskaverse,He2015Kshortest} were introduced to find the shortest path with partial observability of the system. However, these models require network decomposition or partitioning in advance, and solve optimal paths with polynomial-bounded time complexity at best \cite{JOHNSON2016partitioning}. Some other approach heavily relies on V2I communication \cite{Mostafizi2021decentralized}. Considering the massive amount of iterations of trial-and-error in reinforcement learning, none of these decentralized routing methods provide a suitable design for a learning-based framework.
}

To achieve efficient decentralized dynamic routing, we extend Dijkstra's algorithm to update the optimal route based on the updated intra-link travel times. As shown in Algorithm~\ref{alg:ETA_prepopulation}, first a pre-populating process is carried out where a standard Dijkstra's algorithm is run to get the $\mathsf{ETA}^0$ from each intersection to the destination. For each intersection, the next intersection $\mathsf{Next}^0$ along the shortest path is also calculated. For an intersection $i$, the result $\mathsf{ETA}_i^0$ and $\mathsf{Next}_i^0$ are stored locally in the intersection. We assume this process can be done before the EMV leaves the dispatching hub. This is reasonable since a sequence of processes, including call-taker processing, are performed before the EMVs are dispatched. Once the pre-populating process is finished, we can update $\mathsf{ETA}$ and $\mathsf{Next}$ for each intersection efficiently in parallel in a decentralized way, since the update only depends on information of neighboring intersections. 
\begin{figure}[h]
\includegraphics[width=\linewidth]{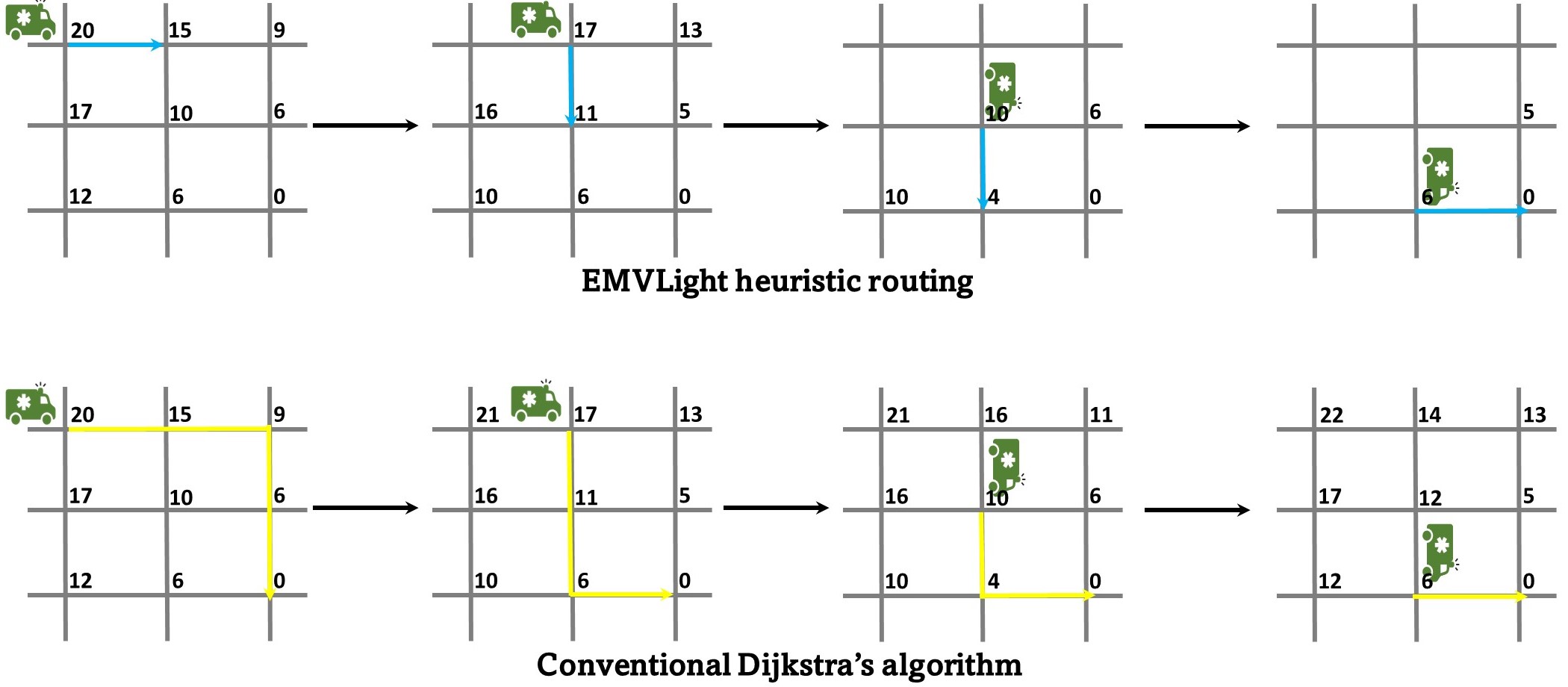}
\centering
\caption{EMVLight routing (top) vs conventional Dijkstra's routing (bottom).}
\label{fig_EMVLight_routing}
\end{figure}
\hs{
Fig. \ref{fig_EMVLight_routing} provides an example comparing between Algorithm.\ref{alg:ETA_prepopulation} and conventional Dijkstra's algorithm on an 3-by-3 traffic network. The numerical value represents the $\mathsf{ETA}$ of each intersection, and it gets updated as described above. EMVLight, rather than solving the full shortest path like the conventional Dijkstra's algorithm, only determines the next link to travel each iteration. A comparison of the worst-case time complexity between the EMVLight routing heuristic and a Dijkstra-based dynamic shortest path method is provided in Table\ref{tab_routing_comparison}. Notice that the Dijkstra' algorithm can be implemented with a Fibonacci heap min-priority queue and solve the shortest path with a time complexity of $\mathcal{O}(|\mathcal{V}|\log{}|\mathcal{V}| + |\mathcal{E}|)$ \cite{Fredman1984Fibonacci}.
\begin{table}[h]
\centering
\fontsize{10.0pt}{10.0pt} \selectfont
\begin{tabular}{@{}ccc@{}}
\toprule
                     & EMVLight heuristic & Dynamic shortest-path based on Dijkstra's \\ \midrule
Initialization & $\mathcal{O}(|\mathcal{V}|\log{}|\mathcal{V}| + |\mathcal{E}|)$    &   -                          \\
updating &   $\mathcal{O}(|\mathcal{V}|)$        &   $\mathcal{O}(|\mathcal{V}|\log{}|\mathcal{V}| + |\mathcal{E}|)$      \\
updating frequency &   $|\mathcal{V}|$  & $M$  \\ \bottomrule
\end{tabular}
\caption{Time complexities of the proposed routing heuristic and the dynamic shortest-path approach. $M$ can be arbitrarily set to determine the updating frequency. The larger $M$ is, the shorter selected route can be.}
\label{tab_routing_comparison}
\end{table}

By adopting the proposed heuristic routing algorithm, we facilitate the RL agent design, which is introduced in Sec. \ref{sec:agent_design}.
}

\begin{algorithm}[t]
    \caption{Decentralized routing for EMVs}
    \label{alg:ETA_prepopulation}
    \SetEndCharOfAlgoLine{}
    \SetKwInOut{Input}{Input}
    \SetKwInOut{Output}{Output}
    \SetKwData{ETA}{ETA}
    \SetKwData{Next}{Next}
    \SetKwFor{ParrallelForEach}{foreach}{do (in parallel)}{endfor}
    \Input{\\\hspace{-3.7em}
        \begin{tabular}[t]{l @{\hspace{3.3em}} l}
        $G=(\mathcal{V}, \mathcal{E})$ & traffic map as a graph \\
        $T^t = [T_{ij}^t]$ & intra-link travel time at time $t$ \\
        $i_d$  & index of the destination
        \end{tabular}
    }
    \Output{\\\hspace{-3.7em}
        \begin{tabular}[t]{l @{\hspace{1.5em}} l}
        $\mathsf{ETA}^t = [\mathsf{ETA}^t_i]$ & ETA of each intersection \\
        $\mathsf{Next}^t = [\mathsf{Next}^t_i]$ & next intersection to go \\
        & from each intersection
        \end{tabular}
    }
    \tcc{pre-populating}
    $\mathsf{ETA}^0, \mathsf{Next}^0$ $=$ \texttt{Dijkstra}$(G, T^0, i_d)$\;
    \tcc{dynamic routing}
    \For{$t = 0 \to T$}{
        \ParrallelForEach{$i \in \mathcal{V}$}{
            $\mathsf{ETA}_i^{t+1} \gets \min_{(i, j)\in \mathcal{E}} (\mathsf{ETA}_j^t + T_{ji}^t)$\;
            $\mathsf{Next}_i^{t+1} \gets \arg\min_{\{j|(i, j)\in \mathcal{E}\}}(\mathsf{ETA}_j^{t} + T_{ji}^t$)\;}}
\end{algorithm}
\begin{remark}
    In static Dijkstra's algorithm, the shortest path is obtained by repeatedly querying the $\mathsf{Next}$ attribute of each node from the origin until we reach the destination. In our dynamic Dijkstra's algorithm, since the shortest path changes, at a intersection $i$, we only care about the immediate next intersection to go to, which is exactly $\mathsf{Next}_i$.
\end{remark}

\subsection{Reinforcement Learning Agent Design}\label{sec:agent_design}
While dynamic routing directs the EMV to the destination, it does not take into account the possible waiting times for red lights at the intersections. Thus, traffic signal pre-emption is also required for the EMV to arrive at the destination in the least amount of time. However, since traditional pre-emption only focuses on reducing the EMV travel time, the average travel time of non-EMVs can increase significantly. Thus, we set up traffic signal control for efficient EMV passage as a decentralized RL problem. In our problem, an RL agent controls the traffic signal phases of an intersection based on local information. Multiple agents coordinate the control signal phases of intersections cooperatively to \textbf{(1)} reduce EMV travel time and \textbf{(2)} reduce the average travel time of non-EMVs. First we design 3 agent types. Then we present agent design and multi-agent interactions.

\subsubsection{Types of agents for EMV passage}
\begin{figure}[h]
    \centering
    \includegraphics[width=\linewidth]{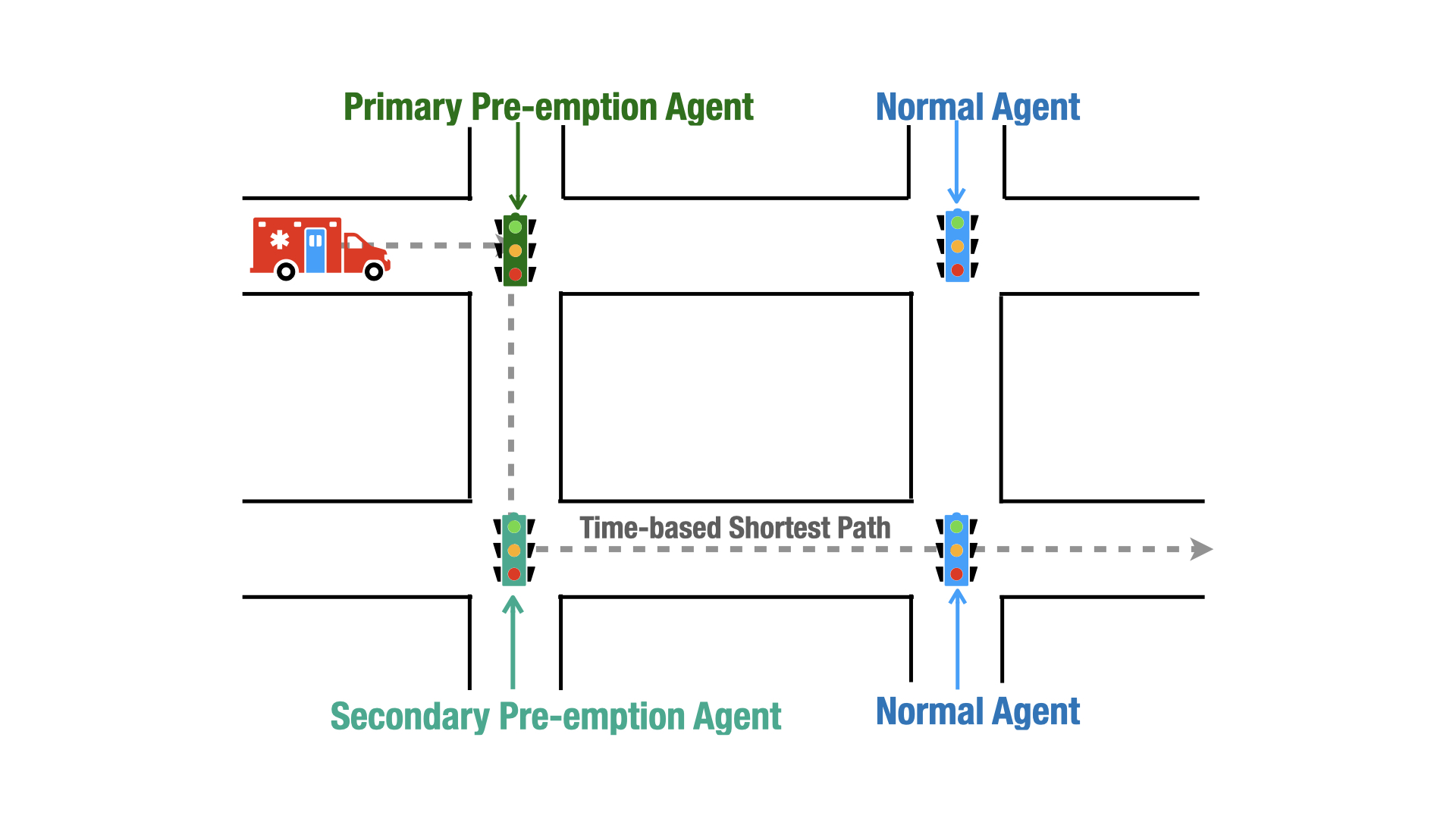}
  \caption{Types of agents through the EMV's passage to the destination.}
  \label{fig_secondary}
\end{figure}

When an EMV is on duty, we distinguish 3 types of traffic control agents based on EMV location and routing (Fig.~\ref{fig_secondary}). An agent is a \emph{primary pre-emption agent} $i_p$ if an EMV is on one of its incoming links. The agent of the next intersection $i_s = \mathsf{Next}_{i_p}$ is refered to as a \emph{secondary pre-emption agent}.
The rest of the agents are \emph{normal agents}. We design these types since different agents have different local goals, which is reflected in their reward designs. 

\subsubsection{Agent design}
\textbf{State}: The state of an agent $i$ at time $t$ is denoted as $s^t_i$ and it includes the number of vehicles on each outgoing lanes and incoming lanes, the distance of the EMV to the intersection, the estimated time of arrival ($\mathsf{ETA}$), and which link the EMV will be routed to ($\mathsf{Next}$), i.e.,
\begin{equation}
    s^t_i = \{x^t(l), x^t(m),  d^t_{\text{EMV}}[L_{ji}], \mathsf{ETA}^{t'}_i, \mathsf{Next}^{t'}_i \},
\end{equation}
where $L_{ji}$ represents the links incoming to intersection  $i$ from its adjacent intersections $j\in\mathcal{N}_{i}$. With a slight abuse of notation, $l$ and $m$ denote the set of incoming and outgoing lanes, respectively. The vector $d^t_{\text{EMV}}$ contains the information about the distance of an EMV to the intersection is an EMV is present. For the intersection shown in Fig.~\ref{fig_movements}, $d^t_{\text{EMV}}$ is a vector of four elements. In particular, for primary pre-emption agents, one of the elements represents the distance of EMV to the intersection in the corresponding link and the rest of the elements are set to -1. For all other agents, $d^t_{\text{EMV}}$ are padded with -1. 

\textbf{Action}: Prior work has focused on using phase switch, phase duration and phase itself as actions. In this work, we define the action of an agent as one of the 8 phases in Fig.~\ref{fig_phases}; this enables more flexible signal patterns as compared to the traditional cyclical patterns. 
Due to safety concerns, once a phase has been initiated, it should remain unchanged for a minimum amount of time, e.g. 5 seconds. Because of this, we set our MDP time step length to be 5 seconds to avoid rapid switch of phases. 

\textbf{Reward}: PressLight has shown that minimizing the pressure is an effective way to encourage efficient vehicle passage. For normal agents, we adopt a similar idea, as shown in Eqn.~\ref{eqn:reward1}. For secondary pre-emption agents, we additionally encourage less vehicles on the link where the EMV is about to enter in order to encourage efficient EMV passage. Thus, the reward is a weighted sum of the pressure and this additional term, with a weight $\beta$, as shown in Eqn.~\ref{eqn:reward2}. For a default setting of balancing EMV navigation and traffic congestion alleviation, we choose $\beta=0.5$. For primary pre-emption agents, we simply assign a unit penalty at each time step to encourage fast EMV passage, as shown in Eqn.~\ref{eqn:reward3}. Thus, depending on the agent type, the local reward for agent $i$ at time $t$ is as follows
\begin{subnumcases}{r_{i}^{t} = \label{eqn:reward}}
  -P_{i}^{t} & $i \notin \{i_p, i_s\}$, \label{eqn:reward1}\\
  - \beta P_{i_s}^{t} - \frac{1-\beta}{|L_{i_pi_s}|}\sum\limits_{l\in L_{i_pi_s}} \frac{x(l)}{x_{max}(l)}  & $i=i_s,$ \label{eqn:reward2}\\
  -1 & $i=i_p$. \label{eqn:reward3} 
\end{subnumcases} 
\textbf{Justification of agent design.} The quantities in local agent state can be obtained at each intersection using various technologies. Numbers of vehicles on each lane $(x^t(l), x^t(m))$ can be obtained by vehicle detection technologies, such as inductive loop \cite{gajda2001vehicle} based on the hardware installed underground. The distance of the EMV to the intersection $d^t_{EMV}[L_{ji}]$ can be obtained by \emph{vehicle-to-infrastructure} technologies such as VANET\cite{buchenscheit2009vanet}, which broadcasts the real-time position of a vehicle to an intersection. Prior work by \citet{wang2013design} and  \citet{noori2016connected} have explored these technologies for traffic signal pre-emption. 

The dynamic routing algorithm (Algorithm~\ref{alg:ETA_prepopulation}) can provide $(\mathsf{ETA}, \mathsf{Next})$ for each agent at every time step. However, due to the stochastic nature of traffic flows, updating the route too frequently might confuse the EMV driver, since the driver might be instructed a new route, say, every 5 seconds. 
There are many ways to ensure reasonable frequency. One option is to inform the driver only once while the EMV travels on a single link. We implement it by updating the state of a RL agent $(\mathsf{ETA}^{t'}_i, \mathsf{Next}^{t'}_i)$ at the time step when the EMV travels through half of a link. For example, if the EMV travels through a link to agent $i$ from time step 11 to 20 in constant speed, then dynamic routing information in $s_i^{16}$ to $s_i^{20}$ are the same, which is $(\mathsf{ETA}_i^{15}, \mathsf{Next}_i^{15})$, i.e., $t'=15$.

As for the reward design, one might wonder how an agent can know its type. As we assume an agent can observe the state of its neighbors, agent type can be inferred from the observation. This will become clearer in Section~\ref{sec:MA2C}.

\subsection{Multi-agent Advantage Actor-critic}
\label{sec:MA2C}
We adopt a multi-agent advantage actor-critic (MA2C) framework similar to \citet{chu2019multi} to address the coupling of EMV navigating and traffic signal control simultaneously in a decentralized manner. The difference is that our local state includes dynamic routing information and our local reward encourages efficient passage of EMV. Here we briefly introduce the MA2C framework.

In a multi-agent network $G(\mathcal{V}, \mathcal{E})$, the neighborhood of agent $i$ is denoted as $\mathcal{N}_i = \{ j | ji\in \mathcal{E} \textrm{ or } ij\in \mathcal{E}\}$. The local region of agent $i$ is $\mathcal{V}_i = \mathcal{N}_i \cup i$. We define the distance between two agents $d(i, j)$ as the minimum number of edges that connect them. For example, $d(i, i) = 0$ and $d(i, j)=1, \forall j \in \mathcal{N}_i$. In MA2C, each agent learns a policy $\pi_{\theta_i}$ (actor) and the corresponding value function $V_{\phi_i}$ (critic), where ${\theta_i}$ and ${\phi_i}$ are learnable neural network parameters of agent $i$.

\textbf{Local Observation.} In an ideal setting, agents can observe the states of every other agent and leverage this global information to make a decision. However, this is not practical in our problem due to communication latency and will cause scalability issues. We assume an agent can observe its own state and the states of its neighbors, i.e., $s^t_{\mathcal{V}_i} = \{s^t_j|j\in \mathcal{V}_i\}$. The agents feed this observation to its policy network $\pi_{\theta_i}$ and value network $V_{\phi_i}$.


\textbf{Fingerprint.} In multi-agent training, each agent treats other agents as part of the environment, but the policy of other agents are changing over time. \citet{foerster2017stabilising} introduce \emph{fingerprints} to inform agents about the changing policies of neighboring agents in multi-agent Q-learning. \citet{chu2019multi} bring fingerprints into MA2C. Here we use the probability simplex of neighboring policies $\pi^{t-1}_{\mathcal{N}_i} = \{\pi^{t-1}_j|j\in \mathcal{N}_i\}$ as fingerprints, and include it into the input of policy network and value network. Thus, our policy network can be written as $\pi_{\theta_i}(a_i^t|s^t_{\mathcal{V}_i}, \pi^{t-1}_{\mathcal{N}_i})$ and value network as $V_{\phi_i}(s^t_{\mathcal{V}_i}, \pi^{t-1}_{\mathcal{N}_i})$, where $s^t_{\mathcal{V}_i}$ is the local observation with spatial discount factor introduced below.

\textbf{Spatial Discount Factor and Adjusted Reward.} MA2C agents cooperatively optimize a global cumulative reward. We assume the global reward is decomposable as $r_t = \sum_{i\in \mathcal{V}} r^t_i$, where $r^t_i$ is defined in Eqn.~\eqref{eqn:reward}. Instead of optimizing the same global reward for every agent, here we employ the spatial discount factor $\alpha$, introduced by \cite{chu2019multi}, to let each agent pay less attention to rewards of agents farther away. The adjusted reward for agent $i$ is 
\begin{equation}
    \Tilde{r}_i^t = \sum_{d=0}^{D_i}\Big( \sum_{j\in\mathcal{V}|d(i, j)=d} (\alpha)^d r^t_j\Big),
\end{equation}
where $D_i$ is the maximum distance of agents in the graph from agent $i$. When $\alpha > 0$, the adjusted reward include global information, it seems this is in contradiction to the local communication assumption. However, since reward is only used for offline training, global reward information is allowed. Once trained, the RL agents can control a traffic signal without relying on global information. 

\textbf{Temporal Discount Factor and Return.} 
The local return $\Tilde{R}^t_i$ is defined as the cumulative adjusted reward $\Tilde{R}^t_i := \sum_{\tau=t}^T \gamma^{\tau-t} \Tilde{r}^\tau_i$, where $\gamma$ is the temporal discount factor and $T$ is the length of an episode. We can estimate the local return using value function,
\begin{equation}
    \Tilde{R}^t_i = \Tilde{r}^t_i + \gamma V_{\phi_i^-}(s^{t+1}_{\mathcal{V}_i}, \pi^{t}_{\mathcal{N}_i}|\pi_{\theta_{-i}^-}),
\end{equation}
where $\phi_i^-$ means parameters $\phi_i$ are frozen and $\theta_{-i}^-$ means the parameters of policy networks of all other agents are frozen. 

\begin{figure}[h]
    \centering
    \includegraphics[width=0.9\linewidth]{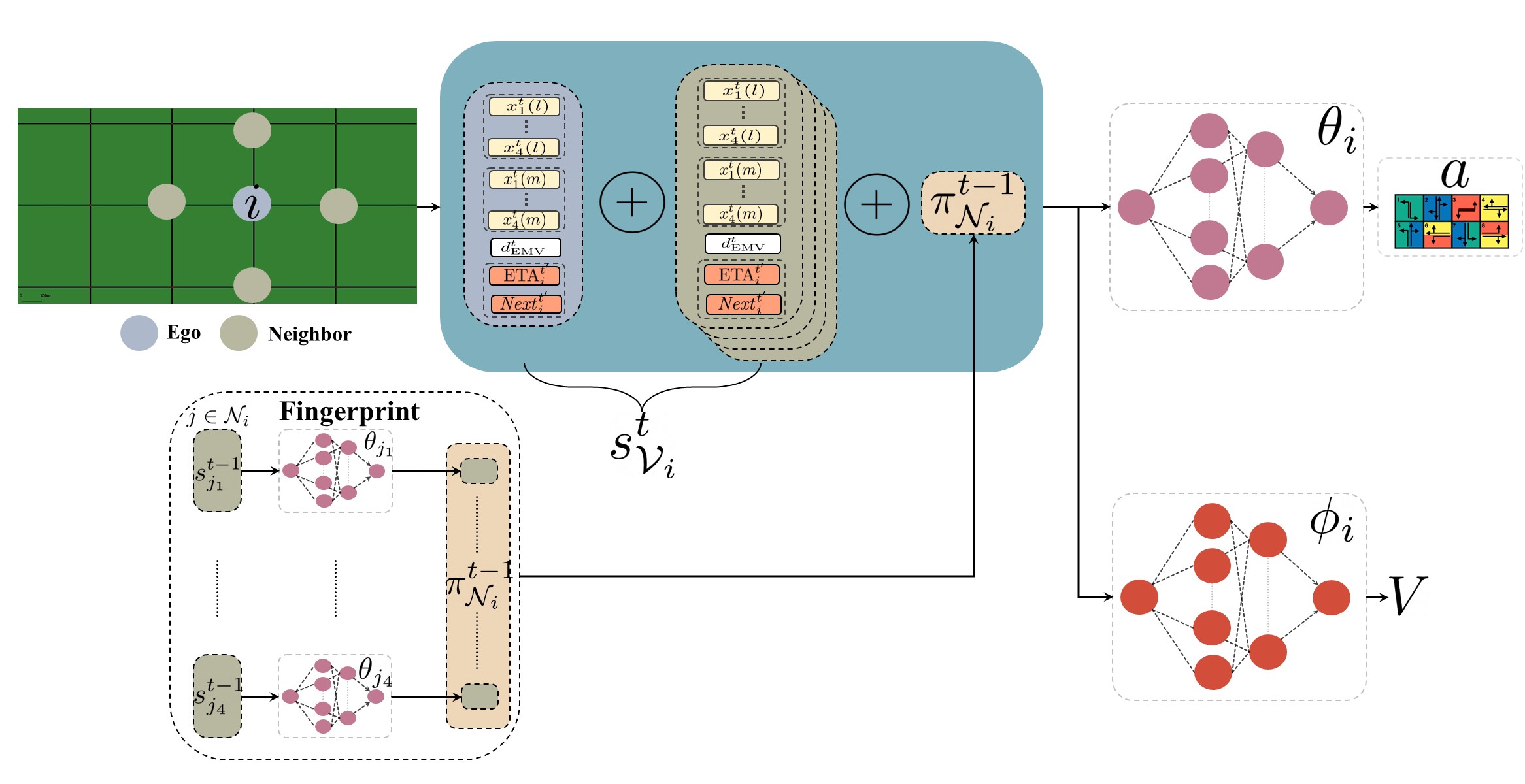}  
    \caption{Overview of MA2C framework for EMVLight's navigation and traffic signal control.}
    \label{fig_model}
\end{figure}

\textbf{Network architecture and training.} 
As traffic flow data are spatial temporal, we leverage a long-short term memory (LSTM) layer along with fully connected (FC) layers for policy network (actor) and value network (critic). Fig. \ref{fig_model} provides an overview for the MA2C frameworks for EMVLight.

\textbf{Value loss function}
With a batch of data $B = \{(s_i^t, \pi_i^t, a_i^t, s_i^{t+1}, r_i^t)_{i\in \mathcal{V}}^{t\in \mathcal{T}}\}$, each agent's value network is trained by minimizing the difference between bootstrapped estimated value and neural network approximated value
\begin{equation}
    \label{eqn:L_v}
    \mathcal{L}_v(\phi_i) = \frac{1}{2|B|} \sum_{B}\Big( \Tilde{R}^t_i - V_{\phi_i}(s^t_{\mathcal{V}_i}, \pi^{t-1}_{\mathcal{N}_i}) \Big)^2.
\end{equation}

\textbf{Policy loss function}
Each agent's policy network is trained by minimizing its policy loss
\begin{align}
    \label{eqn:L_p}
    \mathcal{L}_p(\theta_i) = -& \frac{1}{|B|}\sum_{B} \bigg(\ln \pi_{\theta_i}(a_i^t|s^t_{\mathcal{V}_i}, \pi^{t-1}_{\mathcal{N}_i}) \Tilde{A}^t_i \\
    &- \lambda \sum_{a_i \in \mathcal{A}_i} \pi_{\theta_i} \ln \pi_{\theta_i} (a_i | s^t_{\mathcal{V}_i}, \pi^{t-1}_{\mathcal{N}_i}) \bigg),
\end{align}
where $\Tilde{A}^t_i = \Tilde{R}^t_i - V_{\phi_i^-}(s^t_{\mathcal{V}_i}, \pi^{t-1}_{\mathcal{N}_i})$ is the estimated advantage which measures how much better the action $a^t_i$ is as compared to the average performance of the policy $\pi_{\theta_i}$ in the state $s_i^t$. The second term is a regularization term that encourage initial exploration, where $\mathcal{A}_i$ is the action set of agent $i$. For an intersection as shown in Fig. 1, $\mathcal{A}_i$ contains 8 traffic signal phases.

\textbf{Training algorithm}
Algorithm \ref{alg:training} shows the multi-agent A2C training process. %
\setcounter{algocf}{0}
\renewcommand{\thealgocf}{S\arabic{algocf}}
\begin{algorithm}[ht]
    \caption{Multi-agent A2C Training}
    \label{alg:training}
    \SetEndCharOfAlgoLine{}
    \SetKwInOut{Input}{Input}
    \SetKwInOut{Output}{Output}
    \SetKwData{ETA}{ETA}
    \SetKwData{Next}{Next}
    \SetKwFor{ParrallelForEach}{foreach}{do (in parallel)}{endfor}
    \Input{\\\hspace{-3.7em}
        \begin{tabular}[t]{l @{\hspace{3.3em}} l}
        $T$ & maximum time step of an episode \\
        $N_{\mathrm{bs}}$ & batch size \\
        $\eta_\theta$  & learning rate for policy networks \\
        $\eta_\phi$    & learning rate for value networks \\
        $\alpha$       & spatial discount factor \\
        $\gamma$       & (temporal) discount factor\\ $\lambda$      & regularizer coefficient
        \end{tabular}
    }
    \Output{\\\hspace{-3.7em}
        \begin{tabular}[t]{l @{\hspace{1.4em}} l}
        $\{\phi_i\}_{i\in\mathcal{V}}$ & learned parameters in value networks \\
        $\{\theta_i\}_{i\in\mathcal{V}}$ & learned parameters in policy networks \\
        \end{tabular}
    }
    \textbf{initialize} $\{\phi_i\}_{i\in\mathcal{V}}$, $\{\theta_i\}_{i\in\mathcal{V}}$, $k \gets 0$, $B \gets \varnothing$;
    \textbf{initialize} SUMO, $t \gets 0$, \textbf{get} $\{s^0_i\}_{i\in\mathcal{V}}$\;
    \Repeat{Convergence}{
        \tcc{generate trajectories}
        \ParrallelForEach{$i \in \mathcal{V}$}{
            \textbf{sample} $a^t_i$ from $\pi^t_i$\;
            \textbf{receive} $\Tilde{r}^t_i$ and $s^{t+1}_i$\;
        }
        $B \gets B \cup \{(s_i^t, \pi_i^t, a_i^t, s_i^{t+1}, r_i^t)_{i\in \mathcal{V}}\}$\;
        $t \gets t+1$, $k \gets k+1$\;
        \If{$t == T$}{
            \textbf{initialize} SUMO, $t \gets 0$, \textbf{get} $\{s^0_i\}_{i\in\mathcal{V}}$\;
        }
        \tcc{update actors and critics}
        \If{$k == N_{\mathrm{bs}}$}{
            \ParrallelForEach{$i \in \mathcal{V}$}{
                \textbf{calculate} $\Tilde{r}^t_i$ (Eqn. (4)), $\Tilde{R}^t_i$ (Eqn. (5))\;
                $\phi_i \gets \phi_i - \eta_\phi \nabla \mathcal{L}_v(\phi_i)$\;
                $\theta_i \gets \theta_i - \eta_\theta \nabla \mathcal{L}_p(\theta_i)$\;
            }
            $k \gets 0, B \gets \varnothing$\;
        }
    }
\end{algorithm}

\section{Experimentation}\label{sec_experimentation}
In this section, we demonstrate our RL framework using Simulation of Urban MObility (SUMO) \cite{lopez2018microscopic}
SUMO is an open-source traffic simulator capable of simulating both microscopic and macroscopic traffic dynamics, suitable for capturing the EMV's impact on the regional traffic as well as monitoring the overall traffic flow. An RL-simulator training pipeline is established between the proposed RL framework and SUMO, i.e., the agents collect observations from SUMO and  preferred signal phases are fed back into SUMO.

\subsection{Datasets and Maps Descriptions}
\begin{figure}[h]
    \centering
    \includegraphics[width=0.9\linewidth]{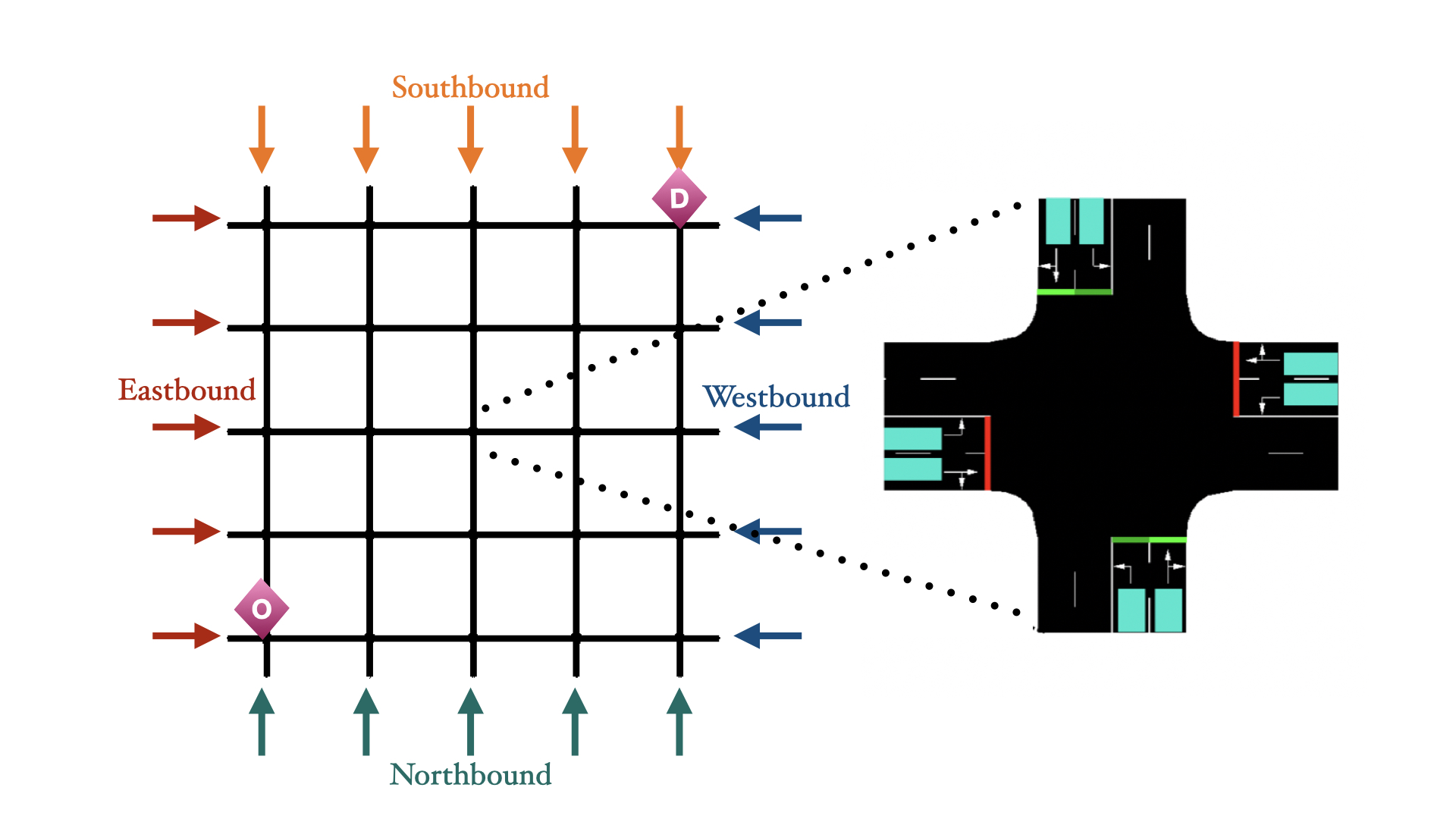}  
    \caption{\emph{Left}: the synthetic $\text{grid}_{5\times 5}$. Origin and destination for EMV are labeled. \emph{Right}: an intersection illustration in SUMO, the teal area are inductive loop detected area. }
  \label{fig_synthetic_map}
\end{figure}

\begin{figure}[h]
    \centering
    \includegraphics[width=0.9\linewidth]{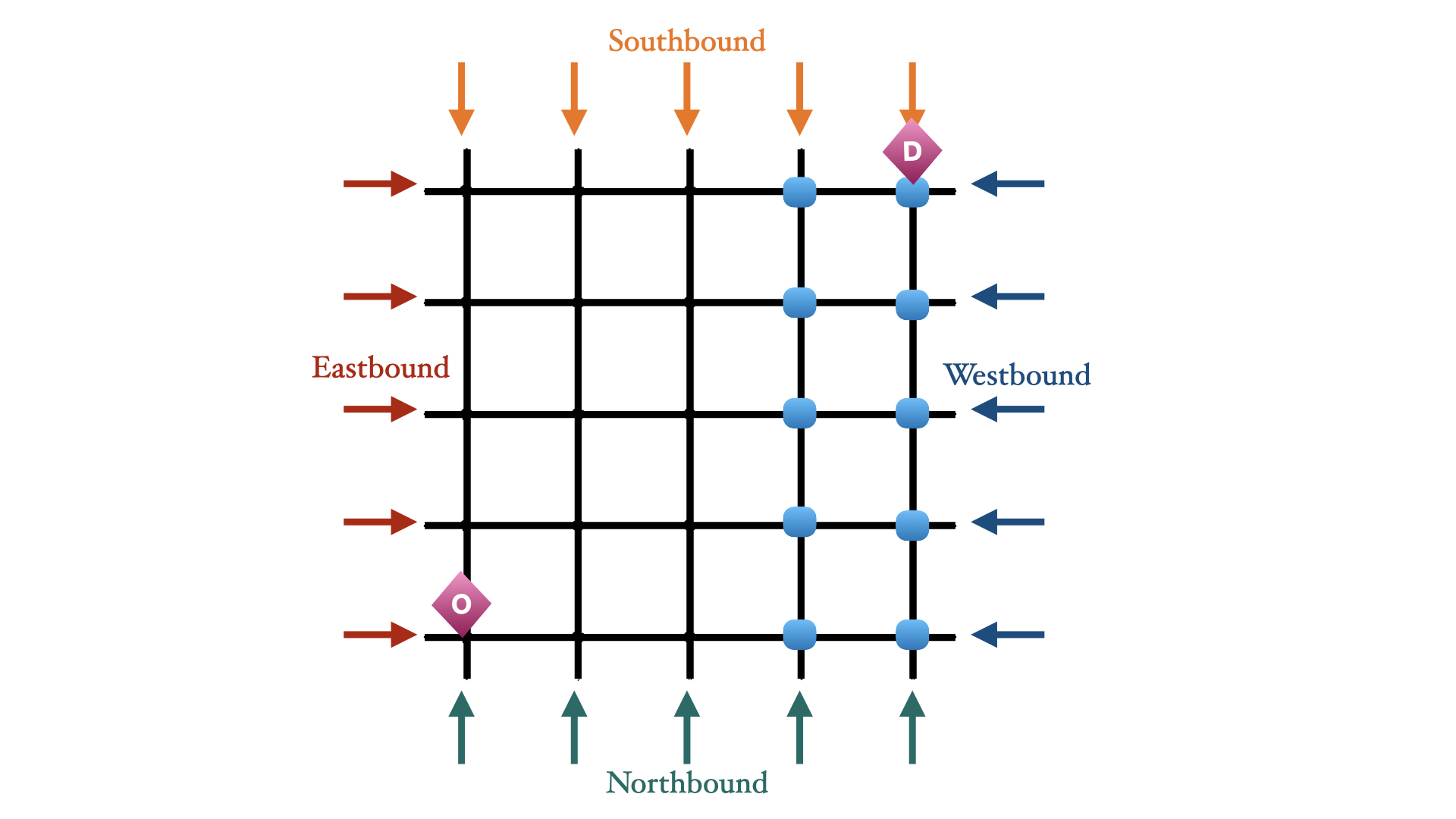}
  \caption{Emergency-capacitated Synthetic $\text{Grid}_{5\times 5}$. Segments towards intersections highlighted by blue have emergency capacities.}
  \label{fig_synthetic_map_reserved}
\end{figure}

We conduct the following experiments based on both synthetic and real-world map. 
\paragraph{Synthetic $\text{Grid}_{5\times 5}$} We synthesize a $5 \times 5$ traffic grid, where intersections are connected with bi-directional links. Each link contains two lanes. We assume all the links have zero emergency capacity. We design 4 configurations of time-varying traffic flows, listed in Table \ref{tab_synthetic_configuration}. 
The origin (O) and destination (D) of the EMV are labelled in Fig.~\ref{fig_synthetic_map}.
The traffic for this map has a time span of 1200s. We dispatch the EMV at $t =600s$ to ensure the roads are compacted when it starts travel.
\begin{table}[h]
\centering
\fontsize{10.0pt}{10.0pt} \selectfont
\begin{tabular}{@{}ccccc@{}}
\toprule[1pt]
\multicolumn{1}{c}{\multirow{2}{*}{Config}} & \multicolumn{2}{l}{Traffic Flow (veh/lane/hr)} & \multirow{2}{*}{Origin}                                                            & \multirow{2}{*}{Destination}                                                     \\ \cmidrule(lr){2-3}
\multicolumn{1}{c}{}                               & Non-peak                 & Peak                &                                                                                    &                                                                                  \\ 
\cmidrule{1-5}
1                                                  & 200                      & 240                 & \multirow{2}{*}{N,S} & \multirow{2}{*}{E,W} \\
\cmidrule{1-3}
2                                                  & 160                      & 320                 &                                                                                    &                                                                                  \\
\cmidrule{1-5}
3                                                  & 200                      & 240                 & \multicolumn{2}{c}{Randomly}                                                                                                               \\
\cmidrule{1-3}
4                                                  & 160                      & 320                 & \multicolumn{2}{c}{generated}     \\ \bottomrule[1pt]                                                                                                                                   
\end{tabular}
\caption{Configuration for Synthetic $\text{Grid}_{5\times 5}$. Peak flow is assigned from 400s to 800s and non-peak flow is assigned out of this period. For Config. 1 and 2, the vehicles enter the grid from North and South, and exit toward East and West.}
\label{tab_synthetic_configuration}
\end{table}
\paragraph{Emergency-capacitated (EC) Synthetic $\text{Grid}_{5\times 5}$}
This map adopts the same network layout as the Synthetic $\text{Grid}_{5\times 5}$ but with emergency-capacitated segments. As shown in Fig. \ref{fig_synthetic_map_reserved}, segments
approaching intersections highlighted by blue are emergency-capacitated with $C^{EC} = 0.2k$, with $k$ represents the normal vehicle capacity of this segment. All other segments are not emergency-capacitated.

\begin{figure}[h]
    \centering
    \includegraphics[width=\linewidth]{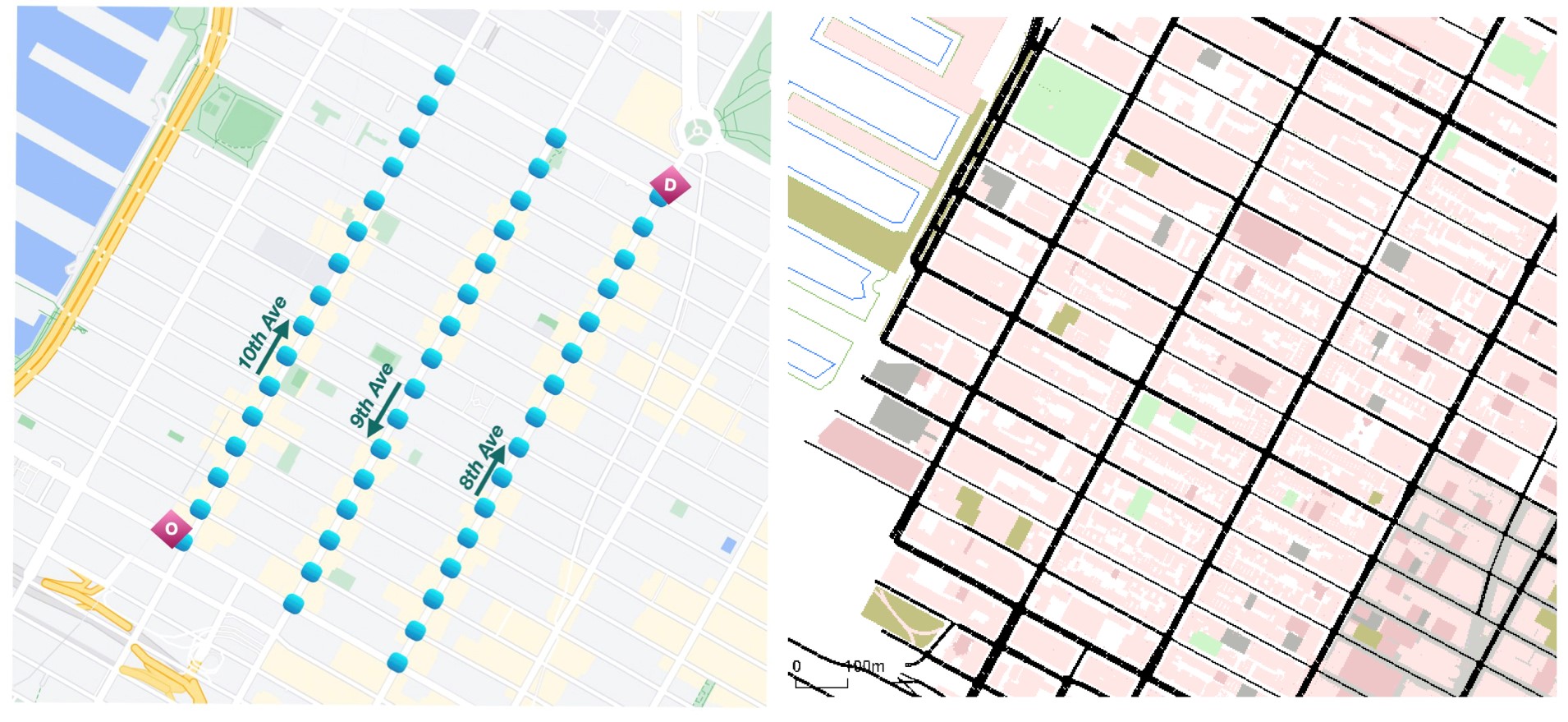}
  \caption{$\textrm{Manhattan}_{16 \times 3}$: a 16-by-3 traffic network in Hell's Kitchen area. Origin and destination for the EMV dispatching are labeled. \emph{Left}: on Google Map; \emph{Right}: in SUMO simulator.}
  \label{fig_manhattan}
\end{figure}

\paragraph{$\text{Manhattan}_{16\times 3}$}
This is a $16 \times 3$ traffic network extracted from Manhattan Hell's Kitchen area (Fig.~\ref{fig_manhattan}) and customized for demonstrating EMV passage. In this traffic network, intersections are connected
by 16 one-directional streets and 3 one-directional avenues. We assume each avenue contains four lanes and each street contains two lanes so that the right-of-way of EMVs and pre-emption can be demonstrated. 
We assume the emergency capacity for avenues and streets are $C^{EC}_{\textrm{avenue}} = 0.2k_{\textrm{avenue}}$ and $C^{EC}_{\textrm{street}} = 0.15k_{\textrm{street}}$, respectively.
The traffic flow for this map is generated from open-source NYC taxi data. Both the map and traffic flow data are publicly available.\footnote{https://traffic-signal-control.github.io/}
The origin and destination of EMV are set to be far away as shown in Fig.~\ref{fig_manhattan}. 


\begin{figure}[h]
    \centering
    \includegraphics[width=\linewidth]{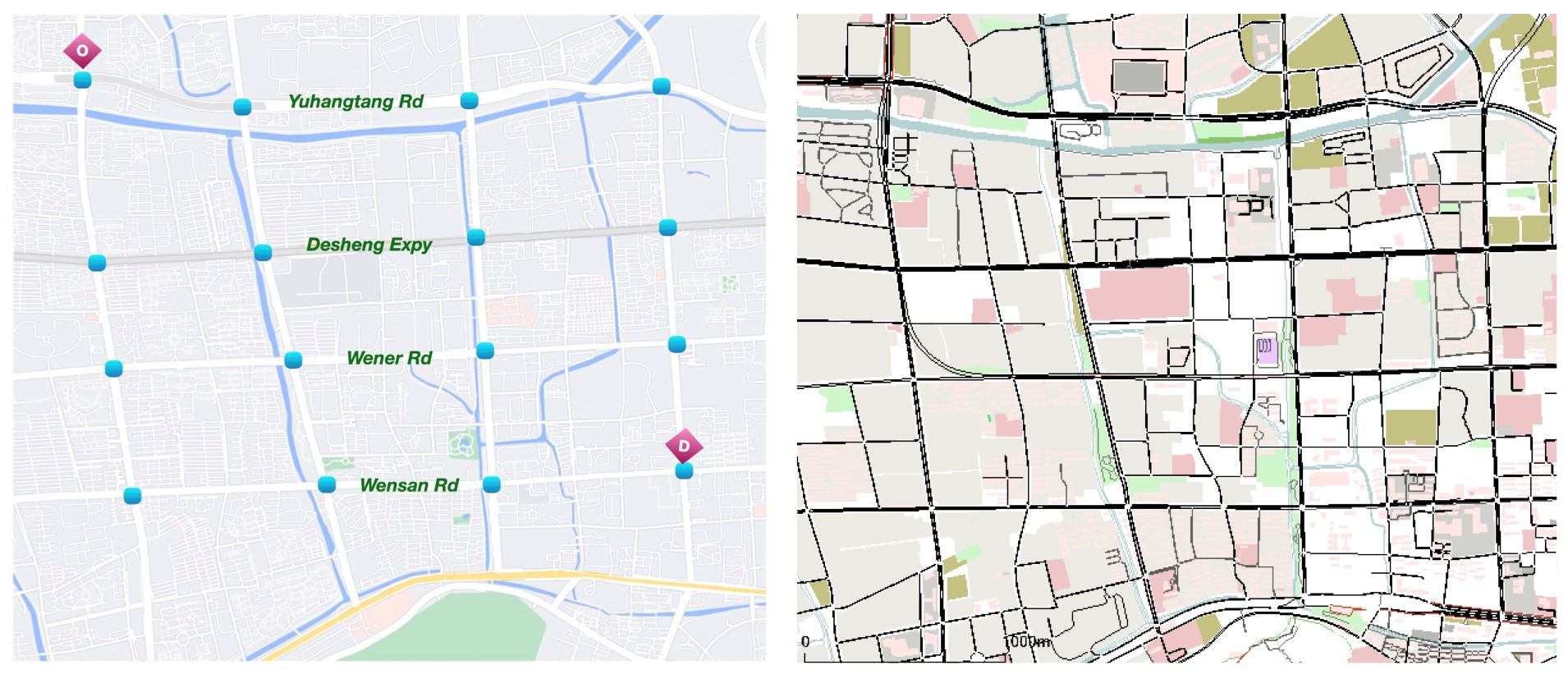}
  \caption{$\textrm{Gudang}_{4 \times 4}$: a 4-by-4 irregular and asymmetric network. Origin and destination for the EMV dispatching are labeled. \emph{Left}: on Google Map; \emph{Right}: in SUMO simulator.}
  \label{fig_gudang}
\end{figure}

\paragraph{$\textrm{Hangzhou}_{4\times 4}$}
An irregular $4 \times 4$ road network represents major avenues in Gudang sub-district in Hangzhou, China. All the road segments are bi-directional with two lanes in each direction. Both the map and traffic flow data are publicly available.
We set the origin and destination for EMV routing as shown in Fig.~\ref{fig_gudang}. The emergency capacity for each segment is set as $C^{EC} = 0.2k$.

\subsection{Benchmark Methods}
Due to the lack of existing RL methods for efficient EMV passage, we select traditional methods and RL methods for each subproblem and combine them to set up benchmarks. 

For traffic signal pre-emption, the most intuitive and widely-used approach is the idea of extending green light period for EMV passage at each intersection which results in a \emph{Green Wave} \cite{corman2009evaluation}. 
\textbf{Walabi (W)} \cite{bieker2019modelling} is 
an effective rule-based method that implemented Green Wave for EMVs in SUMO environment. We integrate Walabi with combinations of routing and traffic signal control strategies introduced below as benchmarks.
We first present two routing benchmarks. 
\begin{itemize}
    \item \textbf{Static routing}: static routing is performed only when EMV starts to travel and the route remains fixed. We adopt A* search as the implementation of static routing since it is a powerful extension to the Dijkstra's shortest path algorithm and is used in many real-time applications. \footnote{Our implementation of A* search employs a Manhattan distance as the heuristic function.}
    \item \textbf{Dynamic routing}: dynamic routing updates the route by taking into account real-time information of traffic conditions. The route is then updated by repeatedly running static routing algorithms. To set up the dynamic routing benchmark, we run A* every 50s as EMV travels. The update interval is set to 50s since running the full A* to update the route is not as efficient as our proposed dynamic Dijkstra's algorithm. 
\end{itemize}

\emph{Traffic signal control benchmarks:}
\begin{itemize}
    \item \textbf{Fixed Time (FT)}: Cyclical fixed time traffic phases with random offset \cite{roess2004traffic} is a policy that split all phases with an predefined green ratio. The coordination between traffic signals are predefined so it is not updated based on real-time traffic. Because of its simplicity, it is the default strategy in real traffic signal control for steady traffic flow.
    \item \textbf{Max Pressure (MP)}: \citet{varaiya2013max} studies max pressure control and use it as the main criterion for selecting traffic signal phases. It defines pressure for each signal phases and aggressively selects the traffic signal phase with maximum pressure to smooth congestion. Hence the name Max Pressure. It is the state-of-the-art network-level signal control strategy that is not based on learning.
    \item \textbf{Coordinated Deep Reinforcement Learners (CDRL)}: CDRL \cite{van2016coordinated} is a Q-learning based coordinator which directly learns joint local value functions for adjacent intersections. It extends Q-learning from single-agent scenarios to multi-agent scenarios.
    It also employs transfer planning and max-plus coordination strategies for joint intersection coordination. 
    \item \textbf{PressLight (PL)}: PL \cite{wei2019presslight} is also a Q-learning based method for traffic signal coordination. It aims at optimizing the pressure at each intersection. However, it defines pressure for each intersection, which is slightly different from the definition in Max Pressure. Our definition of pressure Eqn.~\eqref{eqn:reward}. is also different from that in PL.
    \item \textbf{CoLight (CL)}: CoLight \cite{wei2019colight} uses a graph-attentional-network-based reinforcement learning method for large scale traffic signal control. It adjusts queue length with information from neighbor intersections.
\end{itemize}


\subsection{Metrics}
We evaluate performance of all strategies under two metrics: \emph{EMV travel time}, which reflects the routing and pre-emption ability, and \emph{average travel time}, which indicates the ability of traffic signal control for efficient vehicle passage. Vehicles which have completed their trips during the simulation interval are counted when calculating the average travel time.

\section{Results and Discussion}\label{sec_result}

In this section, we demonstrate the performance of EMVLight and compare it against that of all benchmark methods on four experimentation maps. The results show a clear advantage of EMVLight under the two metrics. 
In addition, we illustrate the difference of underlying route selection by EMVLight and benchmark methods. We further conduct ablation studies to investigate the contribution of different components to EMVLight's performance.

\subsection{Performance Comparison}\label{sec:metrics_comparison}
To evaluate the performance of the proposed EMVLight and all benchmark methods, we conduct SUMO simulation  with five independent runs for each setting. Randomly generated seeds are used in learning-based methods. Means as well as standard deviations of the simulation results are reported for a full numerical assessment. The differences in simulation results for the same setting under independent SUMO runs are coming from configuration noise during generation, such as vehicles' lengths/accelerations/lane-changing eagerness, and, for RL-based methods, random seeds for initialization.

We provide implementation details of EMVLight on different experimentation settings in \ref{appendix_a}. Hyper-parameters choices for EMVLight and RL-based benchmarks are provided in \ref{appendix_b}.
\subsubsection{\texorpdfstring{Synthetic $\text{Grid}_{5\times 5}$ results}{Synthetic Grid results}}
\begin{table}[h]
\centering
\fontsize{9.0pt}{10.0pt} \selectfont
\begin{tabular}{@{}ccccc@{}}
\toprule
\multirow{2}{*}{Method}           & \multicolumn{4}{c}{EMV Travel Time $T_{\textrm{EMV}}$ [s]}       \\ \cmidrule(l){2-5} 
                                  & Config 1 & Config 2 & Config 3 & Config 4 \\ \midrule
FT w/o EMV                        &    N/A      &    N/A      &    N/A      &    N/A      \\ \midrule
W + static + FT           &   258.18 $\pm$ 5.32   &  273.32  $\pm$ 9.74       &  256.40 $\pm$ 6.20      &   240.84  $\pm$ 4.43      \\
W + static + MP         &    260.22 $\pm$ 10.87      &    272.40 $\pm$ 10.92       &   265.74 $\pm$ 11.98       &    242.32 $\pm$ 9.48      \\
W + static + CDRL  &    269.42 $\pm$ 7.32      &   282.20 $\pm$ 5.28      &   276.14 $\pm$ 2.58       &   280.32 $\pm$ 4.82       \\
W + static + PL          &   270.68 $\pm$ 9.13       &  279.14 $\pm$ 9.22        &  281.42 $\pm$ 5.62        &   266.10 $\pm$ 8.32       \\
W + static + CL  &    255.72 $\pm$ 4.23      &   272.06 $\pm$ 8.13      &   270.22 $\pm$ 2.81       &  277.12 $\pm$ 6.10       \\
\midrule
W + dynamic + FT          &   229.38 $\pm$ 8.28       &    212.87 $\pm$ 3.17     &   218.46 $\pm$ 4.28       &    220.69 $\pm$ 7s.96      \\
W + dynamic + MP       &   220.48 $\pm$ 9.26      &  208.08 $\pm$ 12.90         &  212.46 $\pm$ 9.82        &   220.98 $\pm$ 10.62       \\
W + dynamic + CDRL &  239.84 $\pm$ 5.24        &   219.15 $\pm$ 8.26       &  211.86 $\pm$ 7.13        &   232.46 $\pm$ 10.16       \\
W + dynamic + PL          &   243.32 $\pm$ 13.86       &   244.82 $\pm$ 10.52       &  250.12 $\pm$ 8.13        &   255.02 $\pm$ 12.76       \\
W + dynamic + CL  &    220.12 $\pm$ 4.19      &  209.12 $\pm$ 4.76      &   224.00 $\pm$ 5.31       &   226.32 $\pm$ 4.13       \\
 \midrule
EMVLight    & \textbf{195.46} $\pm$ 7.48         &  \textbf{190.66} $\pm$ 8.28         &   \textbf{183.12} $\pm$ 6.43        &    \textbf{189.44} $\pm$ 8.32       \\ \bottomrule
\end{tabular}
\caption{EMV travel time in the four configurations of Synthetic $\text{Grid}_{5\times 5}$. Lower value indicates better performance and the lowest values are highlighted in bold.}
\label{tab_synthetic_emv}
\end{table}
\begin{table}[h]
\centering
\fontsize{9.0pt}{10.0pt} \selectfont
\begin{tabular}{@{}ccccc@{}}
\toprule
\multirow{2}{*}{Method}           & \multicolumn{4}{c}{Average Travel Time $T_{\textrm{avg}}$ [s]}       \\ \cmidrule(l){2-5} 
                                  & Config 1 & Config 2 & Config 3 & Config 4 \\ \midrule
FT w/o EMV                        &    353.43 $\pm$ 4.65    &    371.13  $\pm$ 4.58     &    314.25  $\pm$ 2.90    &    334.10 $\pm$ 3.73     \\ 
\midrule
W + static + FT           &   380.42 $\pm$ 13.35  &  395.17 $\pm$ 15.37      &  350.16 $\pm$ 13.66    &   363.90 $\pm$ 15.39     \\
W + static + MP         &  355.10  $\pm$ 15.36      &    362.09  $\pm$ 16.15     &   318.76  $\pm$ 14.90   &     330.69  $\pm$ 15.52   \\
W + static + CDRL  &   559.19 $\pm$ 3.60 & 540.81 $\pm$ 12.04  & 568.13 $\pm$ 13.25      &  568.13 $\pm$ 6.67       \\
W + static + PL  &   369.52 $\pm$ 8.72    &   372.32 $\pm$ 16.05     &   339.18  $\pm$ 7.17     &   339.12  $\pm$ 5.33     \\ 

W + static + CL  &   365.64 $\pm$ 14.08    &   380.13 $\pm$ 8.20     &   328.42  $\pm$ 17.52     &   333.74  $\pm$ 5.76     \\ 
\midrule
W + dynamic + FT          &  380.76  $\pm$ 10.70       &   404.81 $\pm$ 18.76      &   345.09  $\pm$ 11.60     &   358.90  $\pm$  15.27   \\
W + dynamic + MP       &   360.38 $\pm$ 10.31    &  365.10 $\pm$ 8.33       &   327.98  $\pm$ 18.90     &  351.62  $\pm$ 3.79      \\
W + dynamic + CDRL          & 565.38  $\pm$ 16.10      &   544.29  $\pm$ 19.23     &  598.73  $\pm$ 11.01      &    572.22  $\pm$ 13.94    \\
W + dynamic + PL &  373.17  $\pm$ 17.98      &   387.25 $\pm$ 13.98      &  349.12 $\pm$ 16.25     &   330.21  $\pm$ 17.23     \\ 
W + dynamic + CL &  359.14  $\pm$ 9.52      &   370.45 $\pm$ 4.02      &  320.64 $\pm$ 4.10     &   335.27  $\pm$ 7.62     \\ 
\midrule
EMVLight    &  \textbf{335.09}  $\pm$ 4.13    &   \textbf{333.28}  $\pm$ 8.81     & \textbf{307.90}   $\pm$ 3.89       &   \textbf{321.02} $\pm$ 5.87     \\ \bottomrule
\end{tabular}
\caption{Average travel time for all vehicles which have completed trips in the four configurations of Synthetic $\text{Grid}_{5\times 5}$. }
\label{tab_synthetic_avg}
\end{table}

Table\ref{tab_synthetic_emv} and \ref{tab_synthetic_avg} present the experimental results on average EMV travel time and average travel time on Synthetic $\text{Grid}_{5\times 5}$. In terms of EMV travel time $T_{\textrm{EMV}}$, the dynamic routing benchmark performs better than static routing benchmarks. This is expected since dynamic routing considers the time-dependent nature of traffic conditions and update optimal route accordingly. The best learning and non-learning benchmark methods are dynamics routing with CoLight and Max Pressure, respectively. EMVLight further reduces EMV travel time by 16\% on average as compared to dynamic routing benchmarks. This advantage in performance can be attributed to the design of secondary pre-emption agents. This type of agents learns to ``reserve a link" by choosing signal phases that help clear the vehicles in the link to encourage high speed EMV passage (Eqn.~\eqref{eqn:reward}). 

As for average travel time $T_{\textrm{avg}}$, we first notice that the traditional pre-emption technique (W + static + FT) indeed increases the average travel time by around 10\% as compared to a traditional Fix Time strategy without EMV (denoted as ``FT w/o EMV" in Table \ref{tab_synthetic_avg}), thus decreasing the efficiency of vehicle passage. Different traffic signal control strategies have a direct impact on overall efficiency. Fixed Time is designed to handle steady traffic flow. Max Pressure, as a SOTA traditional method, outperforms Fix Time and, surprisingly, nearly outperforms all RL benchmarks in terms of overall efficiency. This shows that pressure is an effective indicator for reducing congestion and this is why we incorporate pressure in our reward design. Coordinate Learner performs the worst probably because its reward is not based on pressure. PressLight doesn't beat Max Pressure because it has a reward design that focuses on smoothing vehicle densities along a major direction, e.g. an arterial. Grid networks with the presence of EMV make PressLight less effective.
CoLight achieves similar performance as Max Pressure and is the best learning benchmark method.
Our EMVLight improves its pressure-based reward design to encourage smoothing vehicle densities of all directions for each intersection. This enable us to achieve an advantage of 7.5\% over our best benchmarks (Max Pressure).

\subsubsection{Emergency-capacitated Synthetic \texorpdfstring{$\text{Grid}_{5\times 5}$}{Grid} results}

\begin{table}[h]
\centering
\fontsize{9.0pt}{10.0pt} \selectfont
\begin{tabular}{@{}ccccc@{}}
\toprule
\multirow{2}{*}{Method}           & \multicolumn{4}{c}{EMV Travel Time $T_{\textrm{EMV}}$ [s]}       \\ \cmidrule(l){2-5} 
                                  & Config 1 & Config 2 & Config 3 & Config 4 \\ \midrule
FT w/o EMV                        &    N/A      &    N/A      &    N/A      &    N/A      \\ \midrule
W + static + FT           &   254.04 $\pm$ 7.42   &  260.18  $\pm$ 12.03       &  252.12 $\pm$ 11.03      &   232.47  $\pm$ 12.23      \\

W + static + MP         &    233.76 $\pm$ 8.05      &    258.60 $\pm$ 9.06       &   252.74 $\pm$ 13.05      &    233.20 $\pm$ 8.96      \\

W + static + CDRL  &    240.10 $\pm$ 8.65      &   266.28 $\pm$ 8.54      &   258.10 $\pm$ 9.27       &   270.43 $\pm$ 6.18       \\

W + static + PressLight           &   265.28 $\pm$ 7.28       &  269.10 $\pm$ 6.65        &  270.18 $\pm$ 8.83        &   259.20 $\pm$ 7.13       \\

W + static + CoLight  &    250.82 $\pm$ 6.73      &   267.08 $\pm$ 10.21      &   266.12 $\pm$ 4.13       &  270.18 $\pm$ 8.12       \\

\midrule

W + dynamic + FT          &   210.28 $\pm$ 8.90       &    206.18 $\pm$ 7.19     &   210.28 $\pm$ 8.81       &    207.64 $\pm$ 10.02      \\

W + dynamic + MP       &   202.28 $\pm$ 8.54      &  203.20 $\pm$ 9.07         &  206.64 $\pm$ 7.98        &   210.86 $\pm$ 8.59       \\

W + dynamic + CDRL &  218.36 $\pm$ 8.12        &   209.28 $\pm$ 7.19       &  208.180$\pm$ 10.54        &   230.22 $\pm$ 9.22       \\

W + dynamic + PressLight          &   270.08 $\pm$ 10.20       &   238.10 $\pm$ 9.22       &  242.10 $\pm$ 6.98        &   248.24 $\pm$ 10.24       \\

W + dynamic + CoLight  &    216.04 $\pm$ 4.91      &  206.12 $\pm$ 6.27      &   219.26 $\pm$ 6.87      &   223.78 $\pm$ 5.10       \\

 \midrule
EMVLight    & \textbf{150.28} $\pm$ 7.48         &  \textbf{158.20} $\pm$ 6.28         &   \textbf{154.28} $\pm$ 4.19        &    \textbf{159.28} $\pm$ 6.03       \\ \bottomrule
\end{tabular}
\caption{EMV travel time in the four configurations of Synthetic $\text{Grid}_{5\times 5}$ with an emergency capacity co-efficient of 0.25.}
\label{tab_synthetic_ec_emv}
\end{table}

Table\ref{tab_synthetic_ec_emv} shows $T_{\textrm{EMV}}$ of all the methods implemented on the  emergency-capacitated synthetic \texorpdfstring{$\text{Grid}_{5\times 5}$}{Grid} map. 
By comparing Table~\ref{tab_synthetic_emv} and Table~\ref{tab_synthetic_ec_emv}, we conclude that the emergency-capacitated map exhibits overall shorter $T_{\textrm{EMV}}$ in all configurations. 
For benchmark methods, the nonzero emergency capacity shorten $T_{\textrm{EMV}}$ by an average of approximately 12 seconds. 
In particular, dynamic routing-based methods benefit more from the additional emergency capacity, resulting in an average reduction of in 16.28 seconds $T_{\textrm{EMV}}$. This is due to the adaptive nature of dynamic navigation. 
By comparing EMVLight and benchmark methods in Table~\ref{tab_synthetic_ec_emv}, we observe that EMVLight reduce $T_{\textrm{EMV}}$ by up to 50 seconds (25\%) as compared to the best benchmark method in all configurations. 
The substantial difference in $T_{\textrm{EMV}}$ reduction between benchmark methods and EMVLight is due to high success rate of emergency lane forming under coordination, which is investigated further in Section~\ref{subsec_route_selection}. 

\begin{table}[h]
\centering
\fontsize{9.0pt}{10.0pt} \selectfont
\begin{tabular}{@{}ccccc@{}}
\toprule
\multirow{2}{*}{Method}           & \multicolumn{4}{c}{Average Travel Time $T_{\textrm{avg}}$ [s]}       \\ \cmidrule(l){2-5} 
                                  & Config 1 & Config 2 & Config 3 & Config 4 \\ \midrule
FT w/o EMV                        &    353.43 $\pm$ 4.65    &    371.13  $\pm$ 4.58     &    314.25  $\pm$ 2.90    &    334.10 $\pm$ 3.73     \\ 
\midrule

W + static + FT           &   395.28 $\pm$ 5.17  &  410.94 $\pm$ 9.16      &  365.82 $\pm$ 6.14    &   379.64 $\pm$ 6.21     \\

W + static + MP         &  370.52  $\pm$ 6.18      &    375.44  $\pm$ 6.48     &   331.62  $\pm$ 5.92   &     345.13  $\pm$ 8.63 \\

W + static + CDRL  &   575.28 $\pm$ 7.76 & 555.62 $\pm$ 10.04  & 574.91 $\pm$ 19.86      &  585.20 $\pm$ 7.53\\

W + static + PL  &   385.28 $\pm$ 12.09    &   380.83 $\pm$ 10.07     &   360.09  $\pm$ 11.62     &   369.72  $\pm$ 18.02     \\ 

W + static + CL  &   382.17 $\pm$ 6.02    &   380.13 $\pm$ 8.21     &   344.19  $\pm$ 16.02     &   352.07  $\pm$ 4.10     \\ 
\midrule

W + dynamic + FT          &  389.12  $\pm$ 18.21       &   411.98 $\pm$ 15.31      &   353.72  $\pm$ 9.09     &   367.74  $\pm$  16.82   \\

W + dynamic + MP       &   370.28 $\pm$ 12.51    &  362.82 $\pm$ 9.05      &   335.10  $\pm$ 9.16     &  360.02  $\pm$ 17.18     \\

W + dynamic + CDRL          & 575.05  $\pm$ 9.67      &   550.92  $\pm$ 14.06     &  609.26 $\pm$ 11.12      &    578.10  $\pm$ 12.09    \\

W + dynamic + PL &  380.29  $\pm$ 6.10      &   395.28 $\pm$ 5.62      &  359.16 $\pm$ 14.07     &   337.26  $\pm$ 4.96     \\ 

W + dynamic + CL &  366.14  $\pm$ 8.21      &   380.74 $\pm$ 15.84      &  330.44 $\pm$ 17.29     &   343.58  $\pm$ 14.27     \\ 
\midrule

EMVLight    &  \textbf{334.96}  $\pm$ 5.52    &   \textbf{336.18}  $\pm$ 17.09     & \textbf{309.10}   $\pm$ 15.56       &   \textbf{323.76} $\pm$ 17.24     \\ \bottomrule
\end{tabular}
\caption{Average travel time for all vehicles which have completed trips in the four configurations of Synthetic $\text{Grid}_{5\times 5}$ with an emergency capacity coefficient of 0.25. }
\label{tab_synthetic_ec_avg}
\end{table}

As the EMV travels faster and more emergency lanes are formed, the average travel time of non-emergency vehicles increases. Table~\ref{tab_synthetic_ec_avg} shows $T_{\textrm{avg}}$ with emergency capacity added. By comparing Table~\ref{tab_synthetic_avg} and Table~\ref{tab_synthetic_ec_avg}, we observe that with added emergency capacity, the average increase in $T_{\textrm{avg}}$ for non-learning-based and learning-based benchmarks are 12.04 seconds and 7.78 seconds, respectively. Learning-based methods lead to a smaller average increase since agents gradually learn to direct non-EMVs, which are interrupted by EMV passages, to resume their trips as soon as possible. As a result, potential congested queues would not be formed on these segments, effectively reducing the overall $T_{\textrm{avg}}$.

EMVLight, surprisingly, manages to achieve nearly no increase in $T_{\textrm{avg}}$ with added emergency capacity, even though more 
emergency lanes are formed as indicated by smaller $T_{\textrm{EMV}}$. This result shows that EMVLight is able to learn a strong traffic signal coordination strategy while navigating EMVs simultaneously.  The proposed multi-class agent design demonstrates EMVLight's capability of addressing the coupled problems of EMV routing and traffic signal control simultaneously. EMVLight manages to prepare segments for incoming EMVs by reducing the number of vehicles on those segments and restores the impacted traffic in a timely manner after the EMV passage.

\subsubsection{\texorpdfstring{$\textrm{Manhattan}_{16 \times 3}$ results}{Manhattan results}} 

\begin{table}[h]
\centering
\fontsize{9.0pt}{10.0pt} \selectfont
\begin{tabular}{ccc}
\hline
\multirow{2}{*}{Method} & \multicolumn{2}{c}{$\textrm{Manhattan}_{16 \times 3}$}               \\ \cline{2-3} 
                        & $T_{\textrm{EMV}}$     & $T_{\textrm{avg}}$     \\ \hline
FT w/o EMV              &        N/A              &  1649.64                    \\ \hline
W + static + FT         &    817.37 $\pm$ 17.40                  &      1816.43 $\pm$ 68.96                \\
W + static + MP         &      686.72    $\pm$ 19.23          &    917.52  $\pm$ 52.16                 \\
W + static + CDRL        &    702.62 $\pm$ 24.29          & 1247.67  $\pm$ 83.47  \\
W + static + PL         &      626.88    $\pm$ 24.82            &      992.06 $\pm$ 47.67                \\
W + static + CL         &      545.26 $\pm$ 30.21                &       855.28 $\pm$ 41.29               \\ \hline
W + dynamic + FT         &     820.54 $\pm$ 28.86                  &    1808.25 $\pm$ 68.04                  \\
W + dynamic + MP         &    632.68  $\pm$ 13.29                  &   921.18  $\pm$ 49.29                   \\
W + dynamic + CDRL        &      680.62  $\pm$ 20.17              &   1262.39 $\pm$ 60.09                   \\
W + dynamic + PL         & 521.42 $\pm$ 27.62 & 977.62 $\pm$ 53.45   \\
W + dynamic + CL         &    501.26 $\pm$ 28.71                  &  862.94 $\pm$ 45.19                  \\ \hline
EMVLight                &     \textbf{292.82} $\pm$ 16.23                  &         \textbf{782.13} $\pm$ 39.31             \\ \hline
\end{tabular}
\caption{$T_{\textrm{EMV}}$ and $T_{\textrm{avg}}$ for $\textrm{Manhattan}_{16 \times 3}$. The average travel time without the presence of EMVs (1649.64) is retrieved from data.}
\label{tab_Manhattan_results}
\end{table}

Table~\ref{tab_Manhattan_results} presents EMV travel time and average travel time of all the methods on the $\textrm{Manhattan}_{16 \times 3}$ map. In terms of $T_{\textrm{EMV}}$, dynamic routing benchmarks in general result in faster EMV passasge, as expected. 
Compared with benchmark methods, EMVLight produces a considerably low average $T_{\textrm{EMV}}$ of 292.82 seconds, which is 38\% faster than the best benchmark (W+static+CL). 
As for $T_{\textrm{avg}}$, We have similar observation as in the synthetic maps that Max Pressure achieves a similar level of performance on reducing congestion as PressLight and beats CDRL by a solid margin of 25\%. 
CoLight stands out among benchmarks regarding both metrics. Particularly, CoLight shortens $T_{avg}$ by approximately one minute than Max Pressure strategies on this map.

\begin{figure}[h!]
    \centering
    \includegraphics[width=\linewidth]{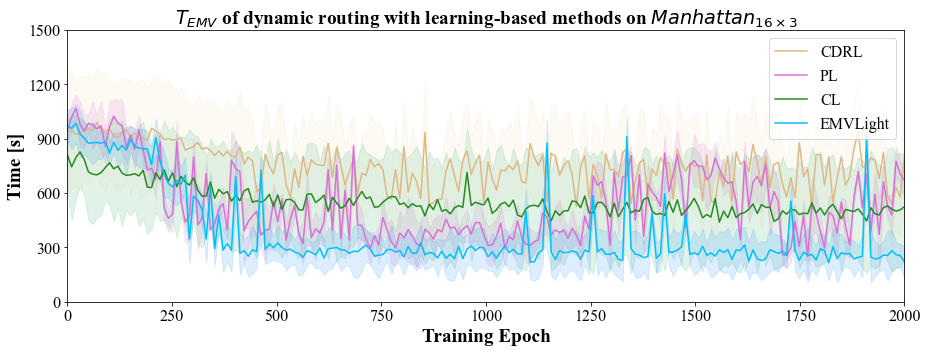}
  \caption{$T_{\textrm{EMV}}$ convergence by learning-based dynamic routing strategies on $\text{Manhattan}_{16\times3}$.}
  \label{fig_emv_learning_curves}
\end{figure}

\begin{figure}[h!]
    \centering
    \includegraphics[width=\linewidth]{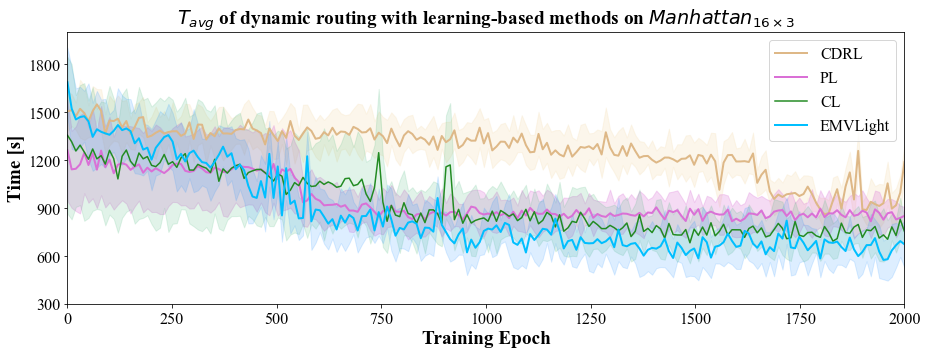}
  \caption{$T_{\textrm{avg}}$ convergence by learning-based dynamic routing strategies on $\text{Manhattan}_{16\times3}$.}
  \label{fig_avg_learning_curves}
\end{figure}

Fig.~\ref{fig_emv_learning_curves} and Fig.~\ref{fig_avg_learning_curves} shows the learning curves of  $T_{\textrm{EMV}}$ and $T_{\textrm{avg}}$, respectively,  in the four RL methods. 
From both figures, we observe that EMVLight has the fastest convergence - in 500 epochs for $T_{\textrm{EMV}}$ and in 1000 epoches for $T_{\textrm{avg}}$ - among all four methods. 
In Fig.~\ref{fig_emv_learning_curves}, both PressLight and CDRL struggle with converging to a stable $T_{\textrm{EMV}}$. 
In Fig.~\ref{fig_avg_learning_curves}, CDRL converges very slowly as compared to the other three methods. 
Both figures shows that CDRL behaves the worst since its DQN design hardly scales with an increasing number of intersections. 
These learning curves demonstrate the fast and stable learning of EMVLight. 

\subsubsection{\texorpdfstring{$\textrm{Hangzhou}_{4 \times 4}$ results}{Hangzhou results}}
\begin{table}[h]
\centering
\fontsize{9.0pt}{10.0pt} \selectfont
\begin{tabular}{ccc}
\hline
\multirow{2}{*}{Method} & \multicolumn{2}{c}{$\textrm{Hangzhou}_{4 \times 4}$}               \\ \cline{2-3} 
                        & $T_{\textrm{EMV}}$     & $T_{\textrm{avg}}$     \\ \hline
FT w/o EMV              &        N/A              &   764.08                   \\ \hline
W + static + FT         & 466.19 $\pm$ 10.25 & 779.13 $\pm$ 12.90 \\
W + static + MP         & 377.20 $\pm$ 14.42 & 404.37 $\pm$ 8.12 \\
W + static + CDRL        & 409.56 $\pm$ 12.06 & 749.10 $\pm$ 10.02 \\
W + static + PL         & 380.82 $\pm$ 6.72 & 425.46 $\pm$ 9.74 \\
W + static + CL         & 368.20 $\pm$ 14.66 & 366.14 $\pm$ 8.25 \\ 
\hline
W + dynamic + FT         & 415.63 $\pm$ 9.03  & 783.89 $\pm$ 10.03\\
W + dynamic + MP         & 328.42 $\pm$ 12.28 & 410.25 $\pm$ 6.23 \\
W + dynamic + CDRL        & 401.08 $\pm$ 15.25 & 755.28 $\pm$ 12.82   \\
W + dynamic + PL         & 321.52 $\pm$ 14.58 & 431.27 $\pm$ 8.24\\
W + dynamic + CL         & 319.84 $\pm$ 11.09 & 370.20 $\pm$ 7.13\\ \hline
EMVLight                & \textbf{194.52} $\pm$ 9.65 & \textbf{331.42} $\pm$ 6.18
\\ \hline
\end{tabular}
\caption{$T_{\textrm{EMV}}$ and $T_{\textrm{avg}}$ for $\textrm{Gudang}_{4 \times 4}$. The average travel time without the presence of EMVs (764.08) is retrieved from data.}
\label{tab_Gudang_results}
\end{table}

Table \ref{tab_Gudang_results} presents $T_{\textrm{EMV}}$ and $T_{\textrm{avg}}$ of EMVLight and benchmark methods on $\textrm{Hangzhou}_{4 \times 4}$. EMVLight achieves the lowest $T_{\textrm{EMV}}$ of 194.52 seconds, \textit{beating} the best benchmark (W+dynamic+CL) by 115 seconds (37\%).
As for $T_{\textrm{avg}}$, EMVLight also has excellent performance, exhibiting a 10\% advantage over W+dynamic+CL, and a 20\% advantage over W+dynamic+MP.
Once again, Max Pressure \textit{outperforms} PressLight in terms of $T_{\textrm{avg}}$. This is consistent with our observations with other maps above, suggesting Max Pressure result in great traffic signal coordination strategies to reduce overall congestion.

Based on results of all introduced experiments, Max Pressure has evinced the consistency to restrict congestion, particularly when restoring stagnant traffic after EMV passages. Consequently, Max Pressure accomplishes the comparable congestion reduction acquirement with, if not better than, RL-based strategies.

\begin{figure}[h]
    \centering
    \includegraphics[width=\linewidth]{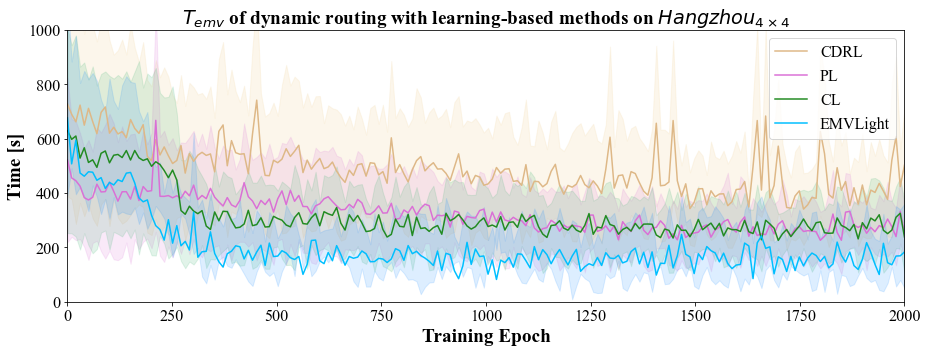}
  \caption{$T_{\textrm{EMV}}$ convergence by learning-based dynamic routing strategies on $\text{Hangzhou}_{4\times4}$.}
  \label{fig_emv_Gudang_learning_curves}
\end{figure}
\begin{figure}[h]
    \centering
    \includegraphics[width=\linewidth]{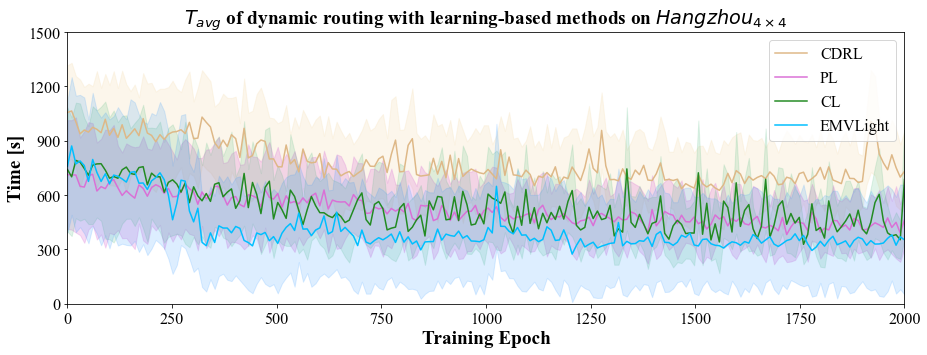}
  \caption{$T_{\textrm{avg}}$ convergence by learning-based dynamic routing strategies on $\text{Gudang}_{4\times4}$.}
  \label{fig_avg_Gudang_learning_curves}
\end{figure}

CDRL fails to learn an effective coordination strategy to manage congestion, where its $T_{\textrm{avg}}$ trivially differs from Fix Time strategy's $T_{\textrm{avg}}$.
CoLight surpasses other methods by shortening $T_{\textrm{avg}}$ by at least 40 seconds but it does not illustrate significant improvement in terms of $T_{\textrm{EMV}}$ from alternative benchmarks.
Learning curves in Fig.~\ref{fig_emv_Gudang_learning_curves} and \ref{fig_avg_Gudang_learning_curves} show that EMVLight converges in 500 epochs. 


\subsection{Routing choices}
\label{subsec_route_selection}
In this section, we investigate the reason behind EMVLight's best performance in $T_{\textrm{EMV}}$ from a routing perspective. In particular, we show how EMVLight is able to leverage emergency capacity to achieve maximum speed passage. 
To understand different route choices between benchmark methods and EMVLight,
we analyze EMV routes on emergency-capacitated Synthetic $\textrm{Grid}_{5\times5}$ as well as $\textrm{Hangzhou}_{4\times4}$ to gain insight into the advantage of EMVLight.

\paragraph{Emergency-capacitated Synthetic \texorpdfstring{$\textrm{Grid}_{5\times5}$}{Grid} routes} 

\begin{figure}
\centering
\subfloat[Config 1]{\label{fig_routing_config1} \includegraphics[width=0.20\textwidth]{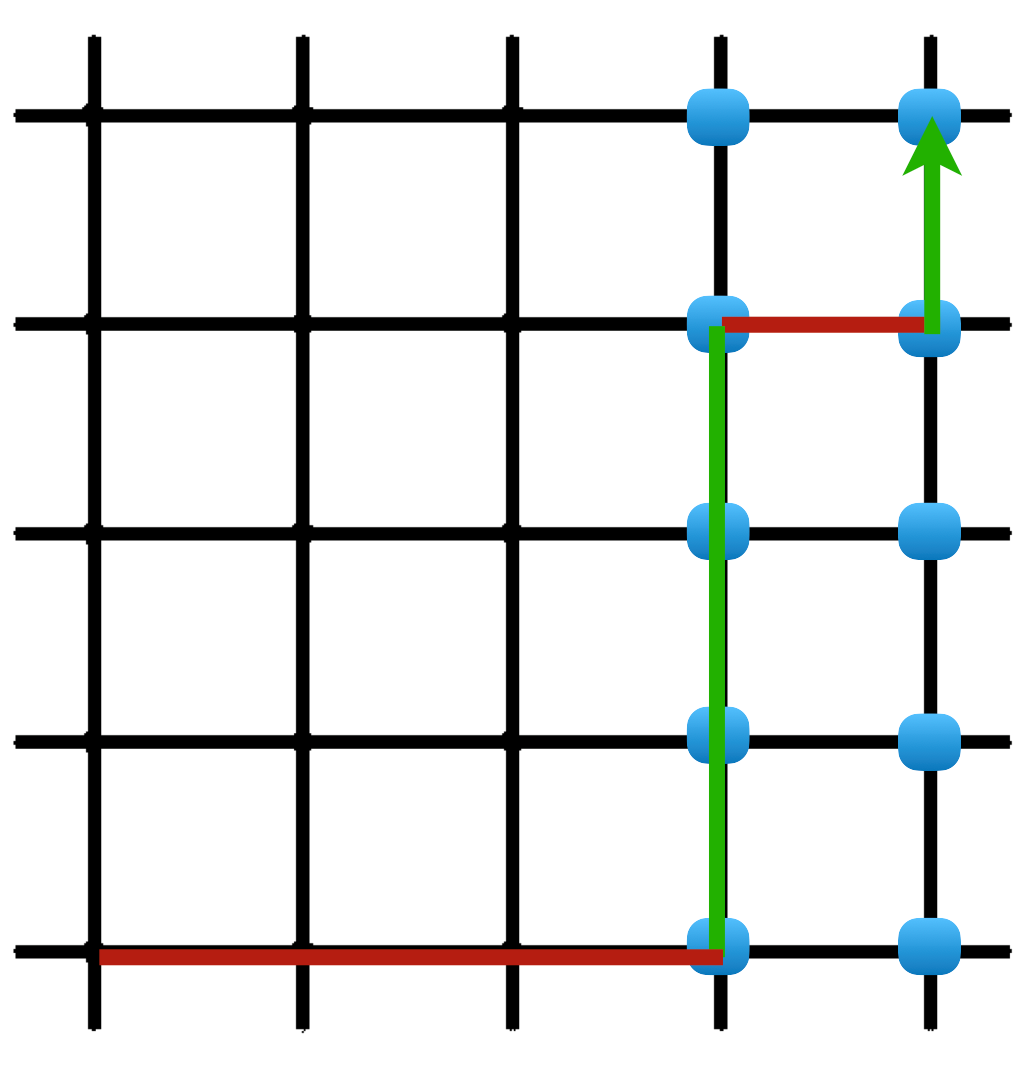}}
\hfill
\subfloat[Config 2]{\label{fig_routing_config2} \includegraphics[width=0.20\textwidth]{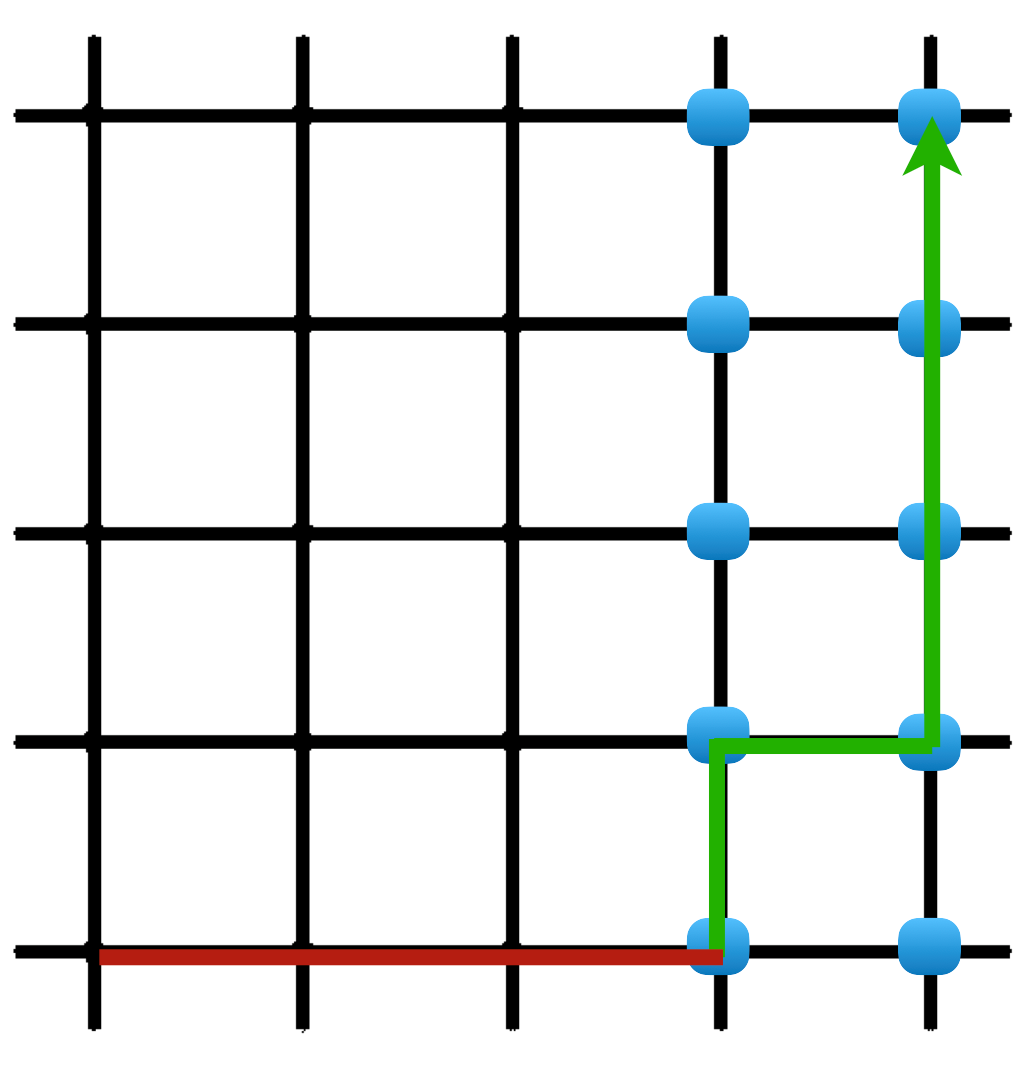}}%
\hfill
\subfloat[Config 3]{\label{fig_routing_config3} \includegraphics[width=0.20\textwidth]{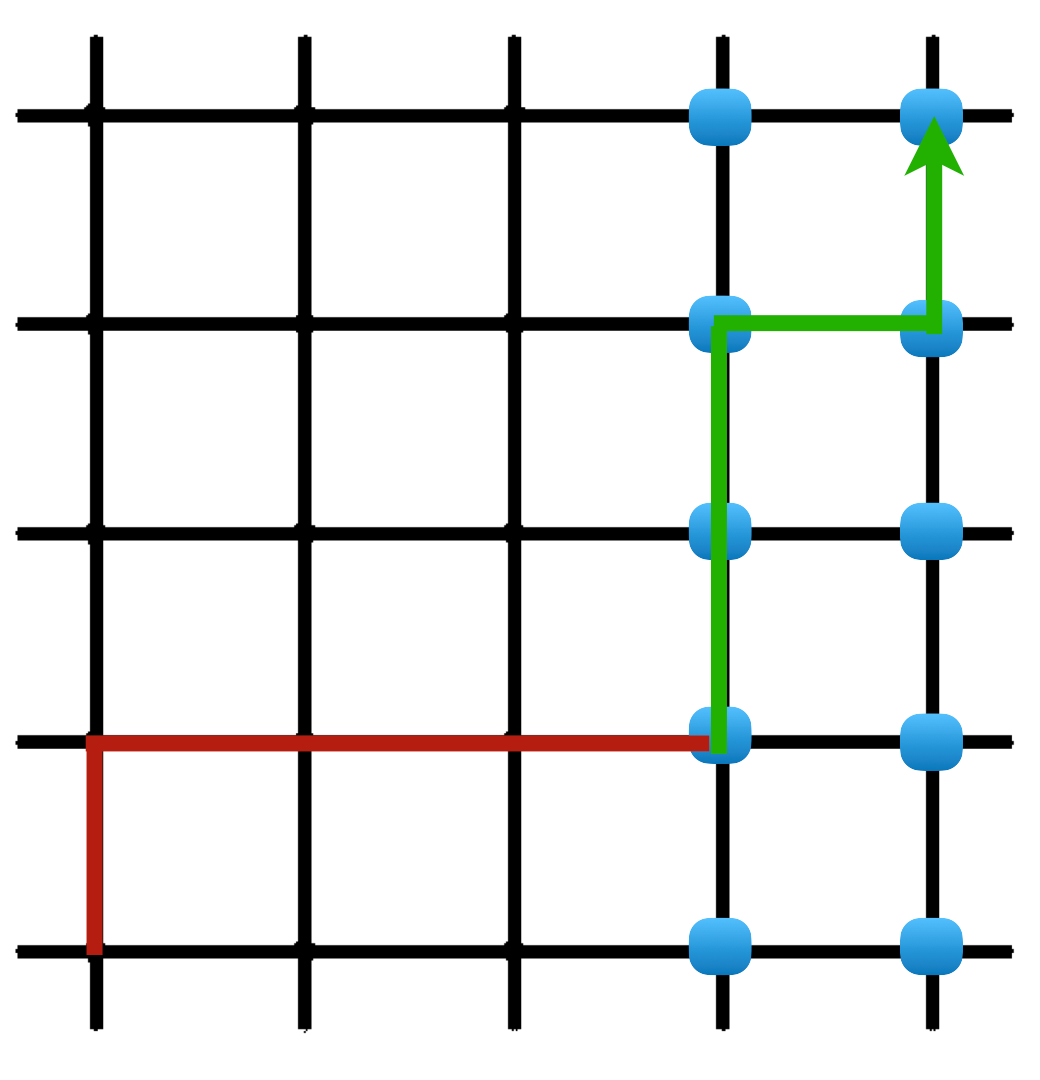}}%
\hfill
\subfloat[config 4]{\label{fig_routing_config4} \includegraphics[width=0.20\textwidth]{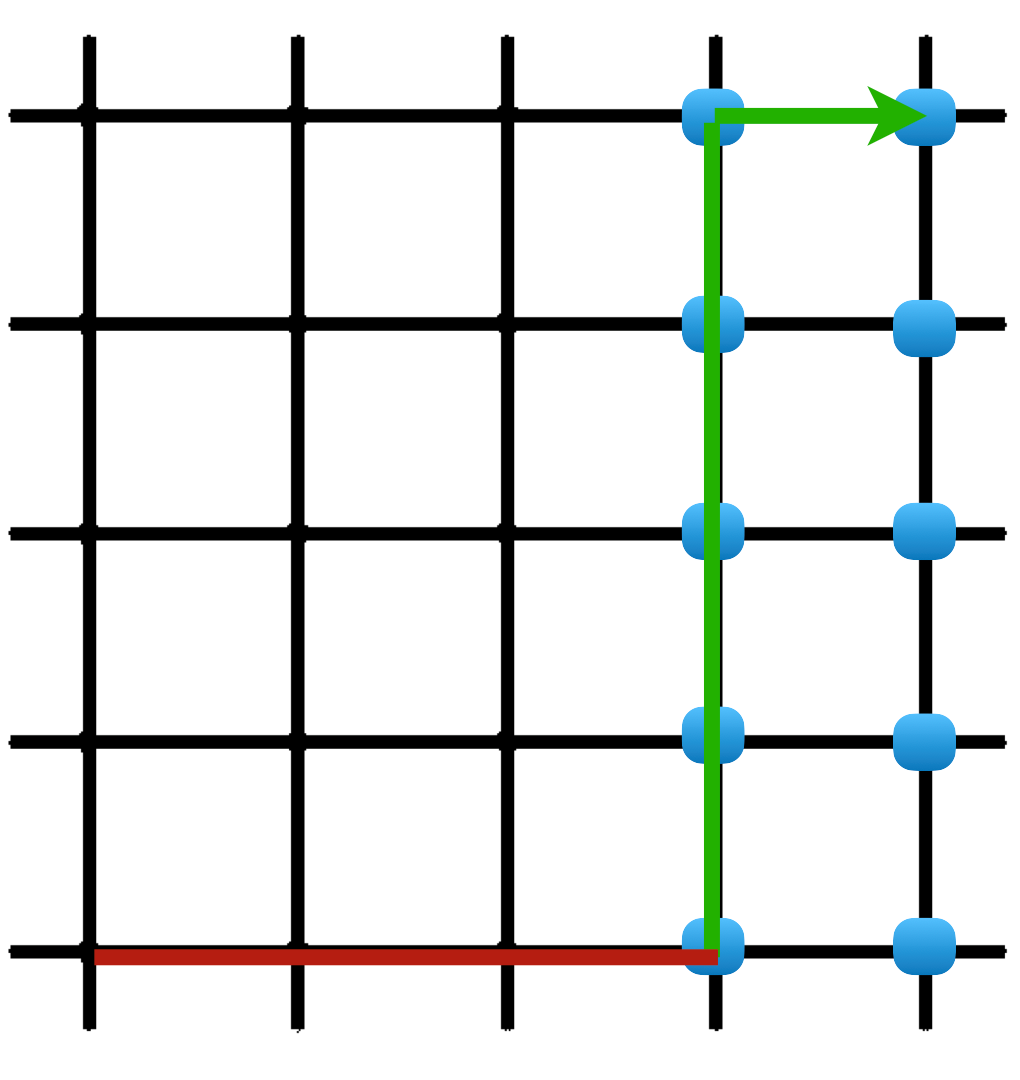}}%
\caption{EMV's routing choice on Emergency-capacitated Synthetic $\text{Grid}_{5\times 5}$ based on EMVLight. Emergency lane established for EMV passage on segments highlighted in green, and not established on segments in red.}
\label{fig_routing_synthetic}
\end{figure}

Routes selected by EMVLight are demonstrated in Fig.~\ref{fig_routing_synthetic} for all four configurations. As this is a regular grid, first we notice that the length of all EMV routes are the Manhattan distance between the origin and destination. This is the shortest length possible to achieve successful EMV dispatch. 
The four routes confirm that EMVLight directs EMVs to enter the region in the east as soon as possible to leverage the extra emergency capacity for full speed passage. This results in 4 full speed links in Config 1 and 3 as well as 5 full speed links in Config 2 and 4.

\begin{figure}[h]
\centering
\begin{subfigure}{.5\textwidth}
  \centering
  \includegraphics[width=.4\linewidth]{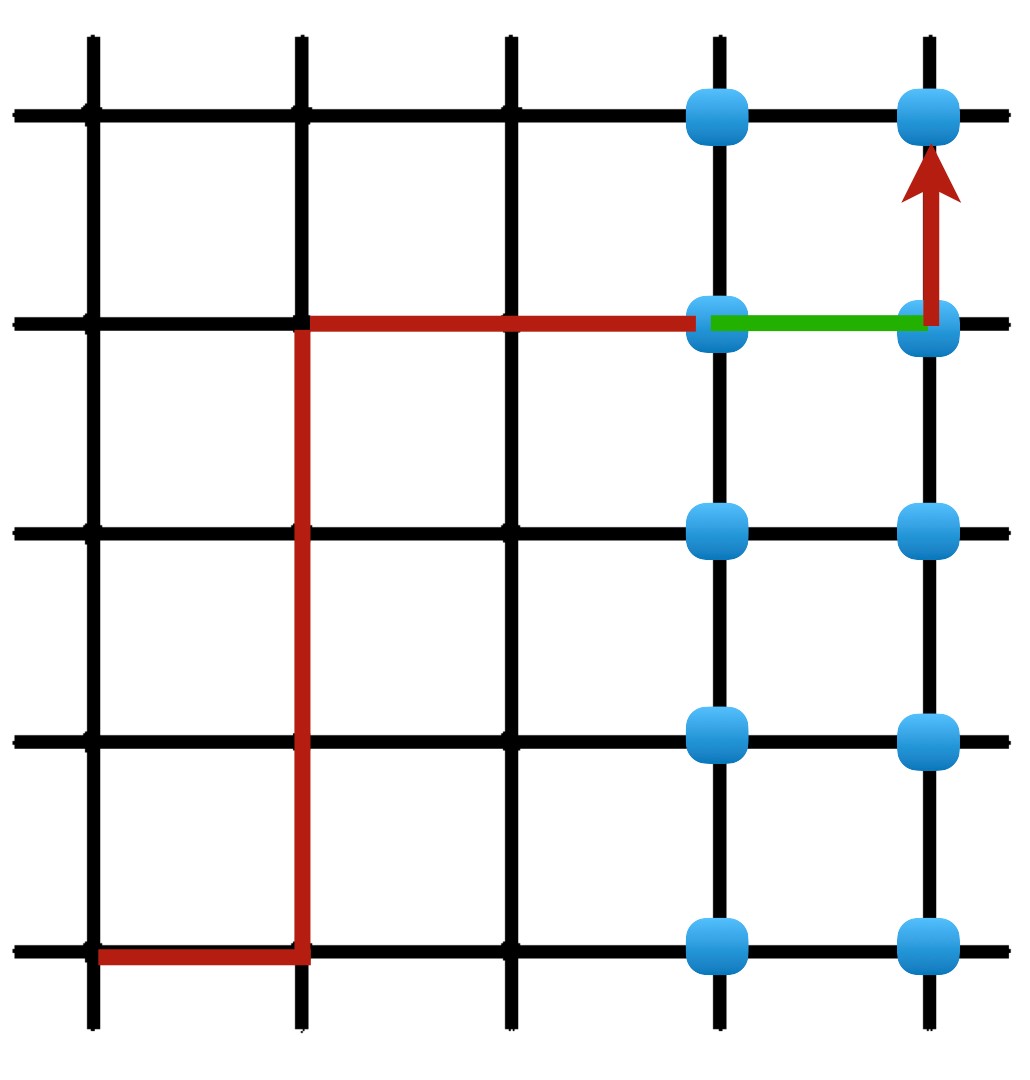}
  \caption{W+static+MP}
  \label{fig_route_choices_static}
\end{subfigure}%
\begin{subfigure}{.5\textwidth}
  \centering
  \includegraphics[width=.4\linewidth]{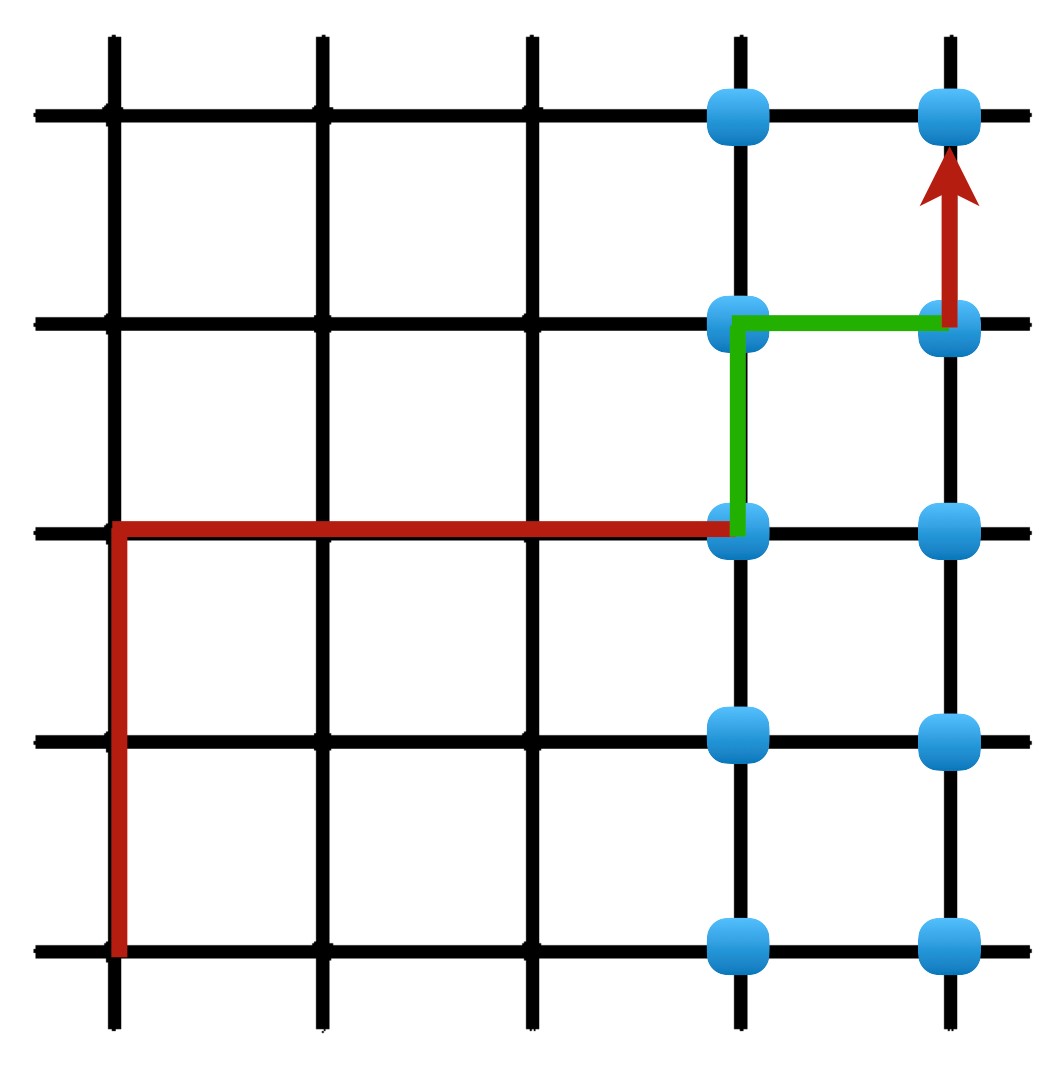}
  \caption{W+dynamic+MP}
  \label{fig_route_choices_dynamics}
\end{subfigure}
\caption{EMV's routing choice on config 1 of Emergency-capacitated Synthetic $\text{Grid}_{5\times 5}$ under MP-based benchmarks.}
\label{fig_route_choices}
\end{figure}

We also present the route choices of W+static+MP and W+dynamic+MP in Config 1, of this map, as shown in Fig.~\ref{fig_route_choices}). 
By comparing these routes with EMVLight (Fig.~\ref{fig_routing_config1}), we can clearly see the benchmark method cannot actively leverage the emergency capacity, generating routes with only 1 full speed segment (Fig.~\ref{fig_route_choices_static}) and 2 full speed segment (Fig.~\ref{fig_route_choices_dynamics}). 


\begin{figure}[ht]
\centering
\begin{subfigure}{.30\textwidth}
    \centering
    \includegraphics[width=.95\linewidth]{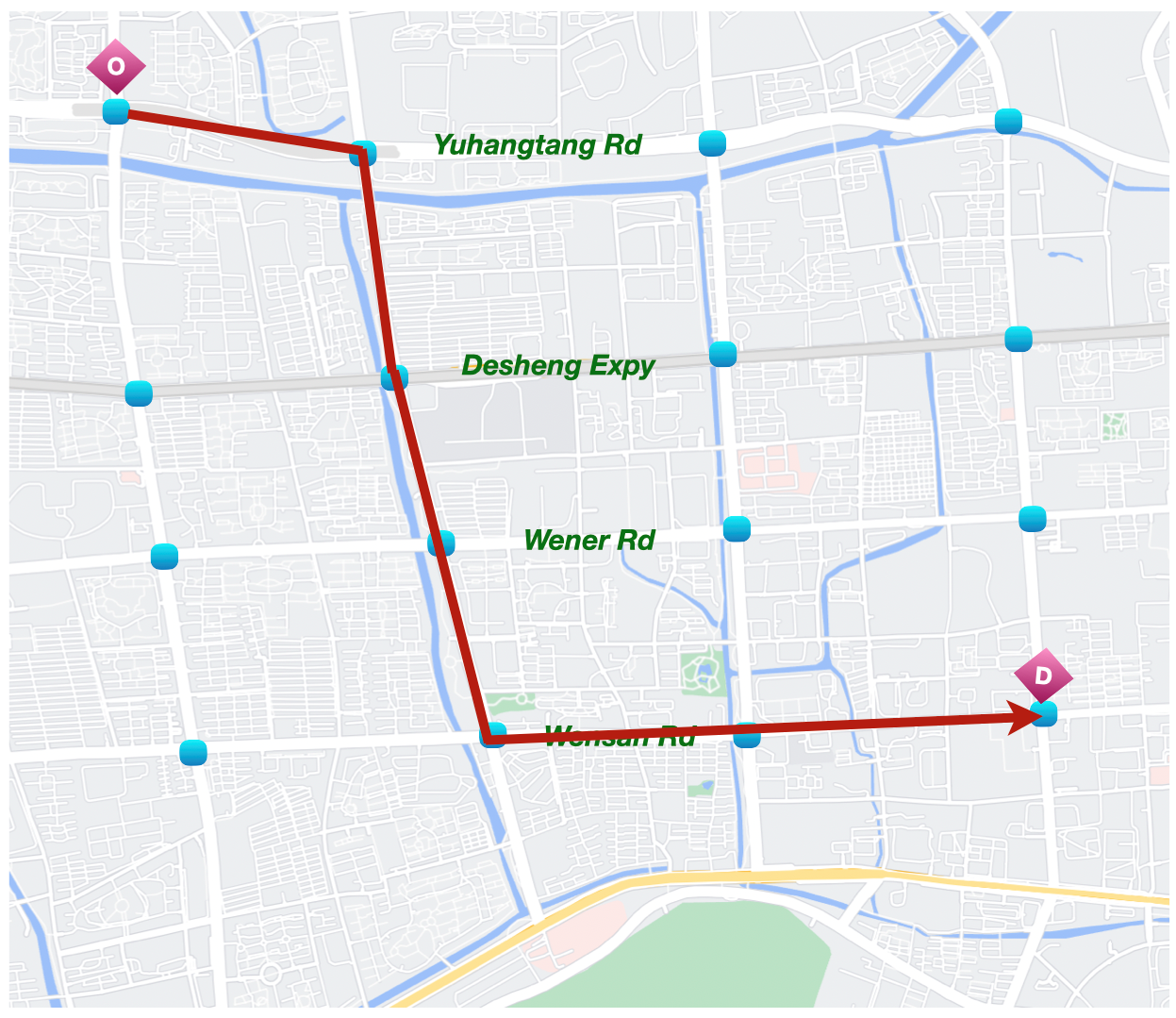}  
    \caption{\bm{$5.5$}km, $277.20 \pm 14.42$s}
    \label{SUBFIGURE LABEL 1}
\end{subfigure}
\begin{subfigure}{.30\textwidth}
    \centering
    \includegraphics[width=.95\linewidth]{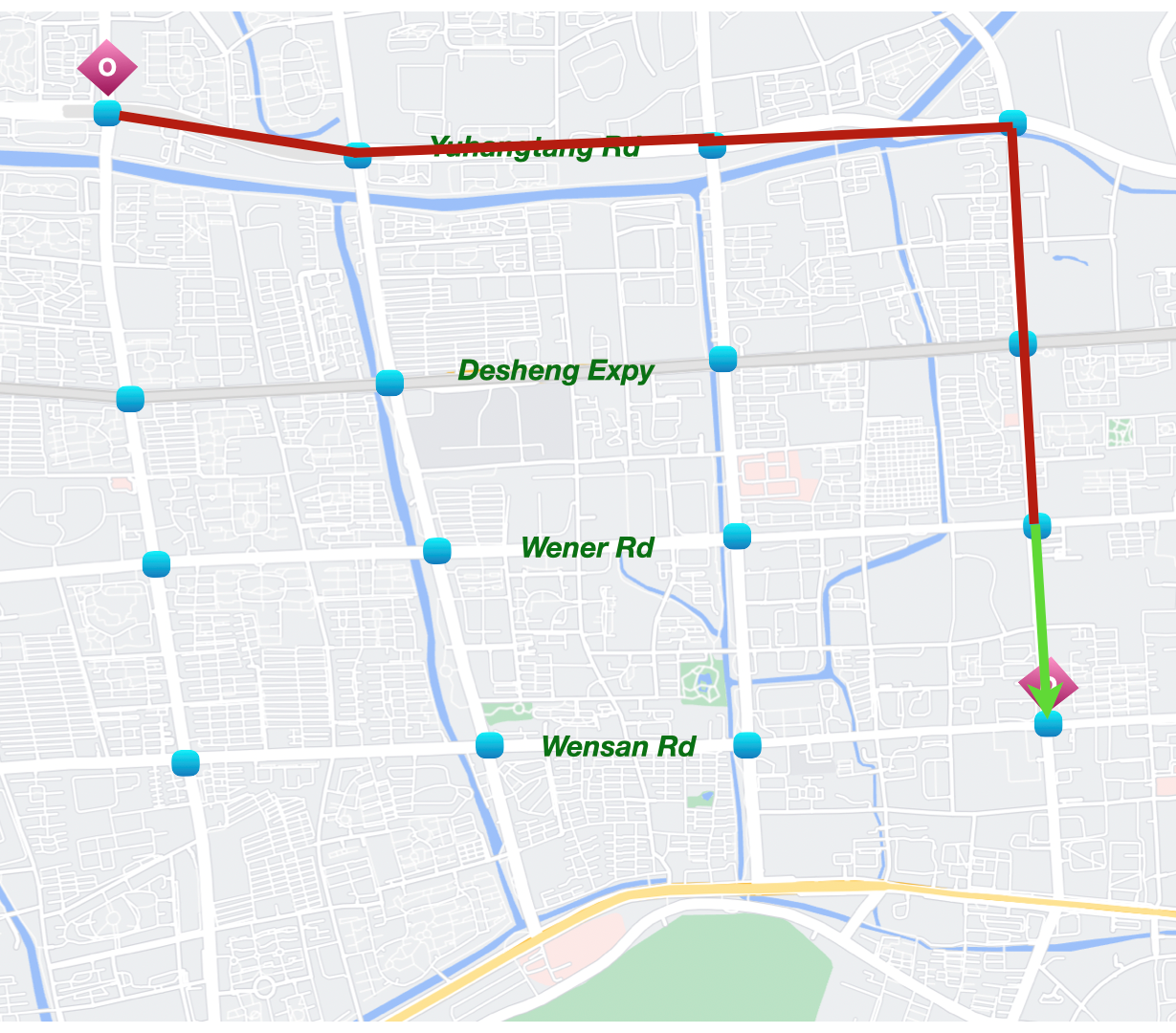}  
    \caption{$5.9$km, $228.42\pm 12.28$s}
    \label{SUBFIGURE LABEL 2}
\end{subfigure}
\begin{subfigure}{.30\textwidth}
    \centering
    \includegraphics[width=.95\linewidth]{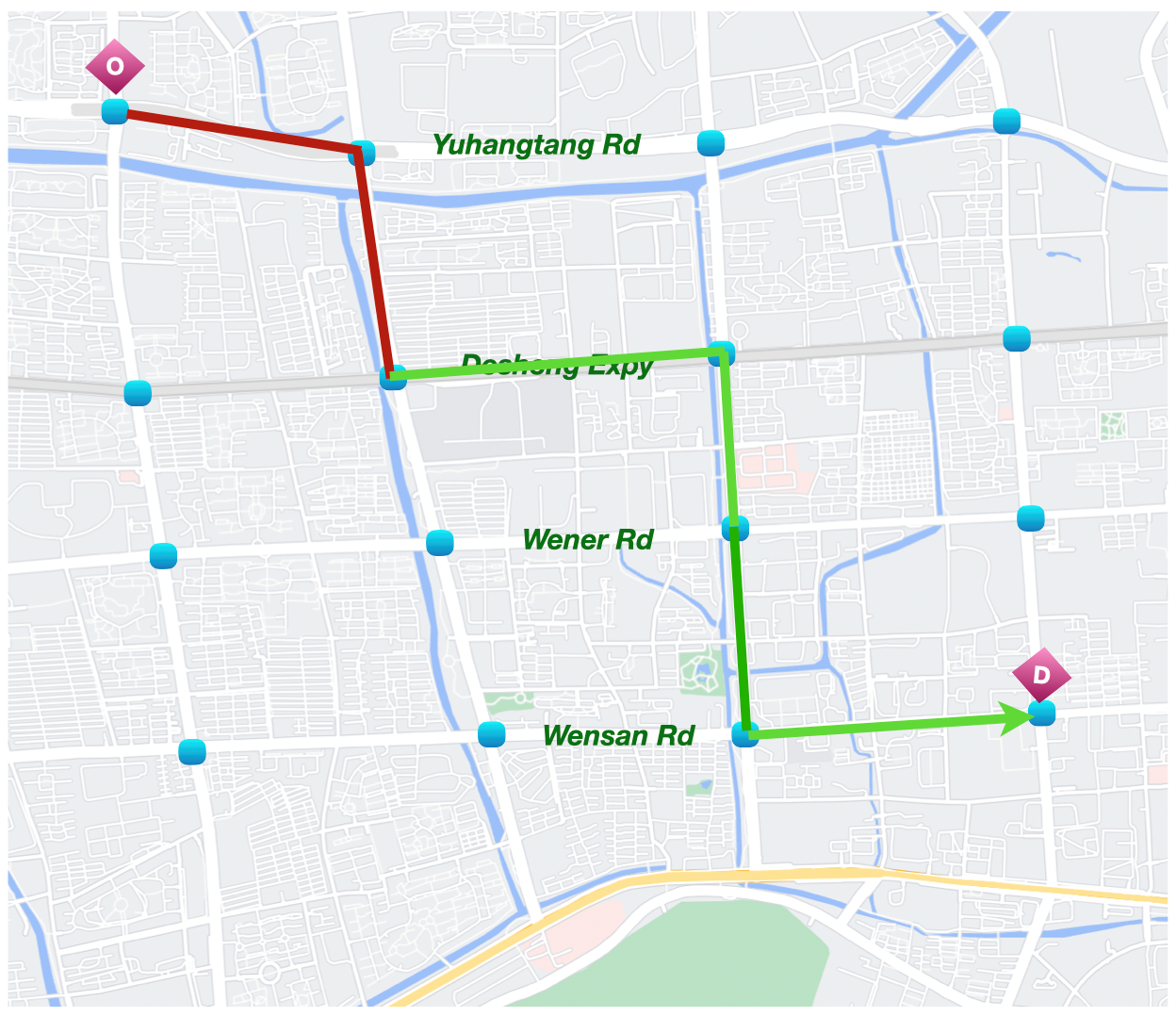}  
    \caption{$6.0$km, $\bm{194.52}\pm9.65$s}
    \label{fig_emvlight}
\end{subfigure}
\caption{The corresponding route selected by (a) W+static+MP, (b) W+dynamic+MP, (c) EMVLight on $\text{Hangzhou}_{4 \times 4}$. Distance and $T_{\textrm{EMV}}$ for the selected path are indicated. The lowest values are highlighted in bold.}
\label{fig_gudang_routings}
\end{figure}

\paragraph{\texorpdfstring{$\textrm{Hangzhou}_{4 \times 4}$}{Hangzhou} routes} 
Since $\textrm{Hangzhou}_{4 \times 4}$ is an irregular grid where different links have different lengths, the routes optimized by different algorithms have different total lengths. This provides another perspective on evaluating routing performance of different models. 
Fig.~\ref{fig_gudang_routings} shows EMV routes given by W+static+MP, W+dynamic+MP and EMVLight in this grid.
By comparing the total distances of the three routes, we find that the route chosen by EMVLight is the longest among the three models. However, EMVLight achieves the smallest EMV travel time on this route. This is because EMVLight is able to coordinate traffic signals to leverage the emergency capacity to let the EMV travel at its maximum speed on more than half of the route, indicated by the green segments. 

As for static routing (W+static+MP), it chooses a fixed route given the traffic conditions upon dispatching, favoring the shortest-in-distance route.
The dynamic routing method (W+dynamic+MP) recalculate the time-based shortest path every 30 seconds as EMV travels. Most of the time, however, Max Pressure fails to reduce the number of vehicles on the upcoming links to enable an emergency lane. Fig.~\ref{SUBFIGURE LABEL 2} shows that the emergency lane is only formed in one of the six links.
EMVLight is able to further reduce EMV travel time partly because the time-based shortest path is updated in real time. More importantly, EMVLight is able to reduce the number of vehicles in upcoming links so that emergency lanes can be formed. This can be attributed to the design of primary and secondary preemption agents in EMVLight, which will be further examined in Sec.\ref{sec_ablation_reward}.


\subsection{Ablation Studies}

\subsubsection{Ablation studies on reward}\label{sec_ablation_reward}
We propose three types of agents and design their rewards (Eqn.~\eqref{eqn:reward}) based on our improved pressure definition and heuristics. 
In order to see how our improved pressure definition and proposed agent types influence the results, we propose three ablation studies:
\begin{enumerate}
    \item replacing our pressure definition by that defined in PressLight
    \item replacing secondary pre-emption agents with normal agents
    \item replacing primary pre-emption agents with normal agents
\end{enumerate}

\begin{table}[h]
\centering
\fontsize{9.0pt}{10.0pt} \selectfont
\begin{tabular}{@{}cccc|c@{}}
\toprule[1pt]
Ablations                        & Ablation 1 & Ablation 2 & Ablation 3 & EMVLight\\ \midrule
$T_{\text{EMV}}$ [s]         & 205.20 $\pm$ 6.92       & 311.52   $\pm$ 5.18                   & 384.71 $\pm$ 8.52    & \textbf{194.52} $\pm$ 9.65       \\
$T_{\text{avg}}$ [s]  & 389.14 $\pm$ 8.40   & 442.73 $\pm$ 6.65 & 444.15 $\pm$ 7.02  & \textbf{331.42} $\pm$ 6.18      \\ \bottomrule[1pt]
\end{tabular}
\caption{Ablation studies on pressure-based reward design and agent types. Experiments are conducted on $\textrm{Hangzhou}_{5 \times 5}$. The lowest value are highlighted in bold.}
\label{tab_ablation_reward}
\end{table}

\begin{figure}[h]
\centering
\begin{subfigure}{.5\textwidth}
  \centering
  \includegraphics[width=.6\linewidth]{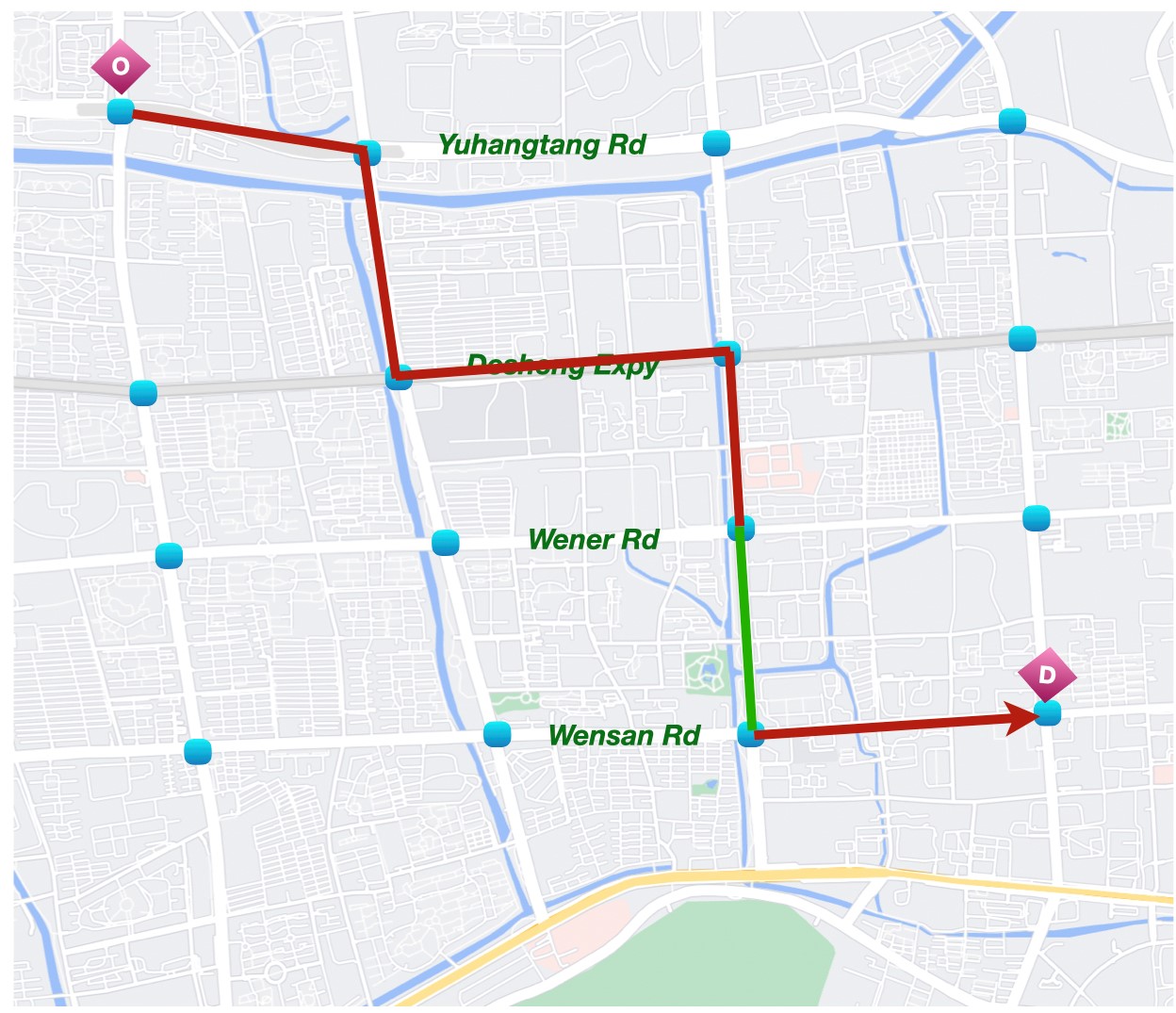}
  \caption{Without secondary agents}
  \label{fig_route_choices_ablation_secondary}
\end{subfigure}%
\begin{subfigure}{.5\textwidth}
  \centering
  \includegraphics[width=.6\linewidth]{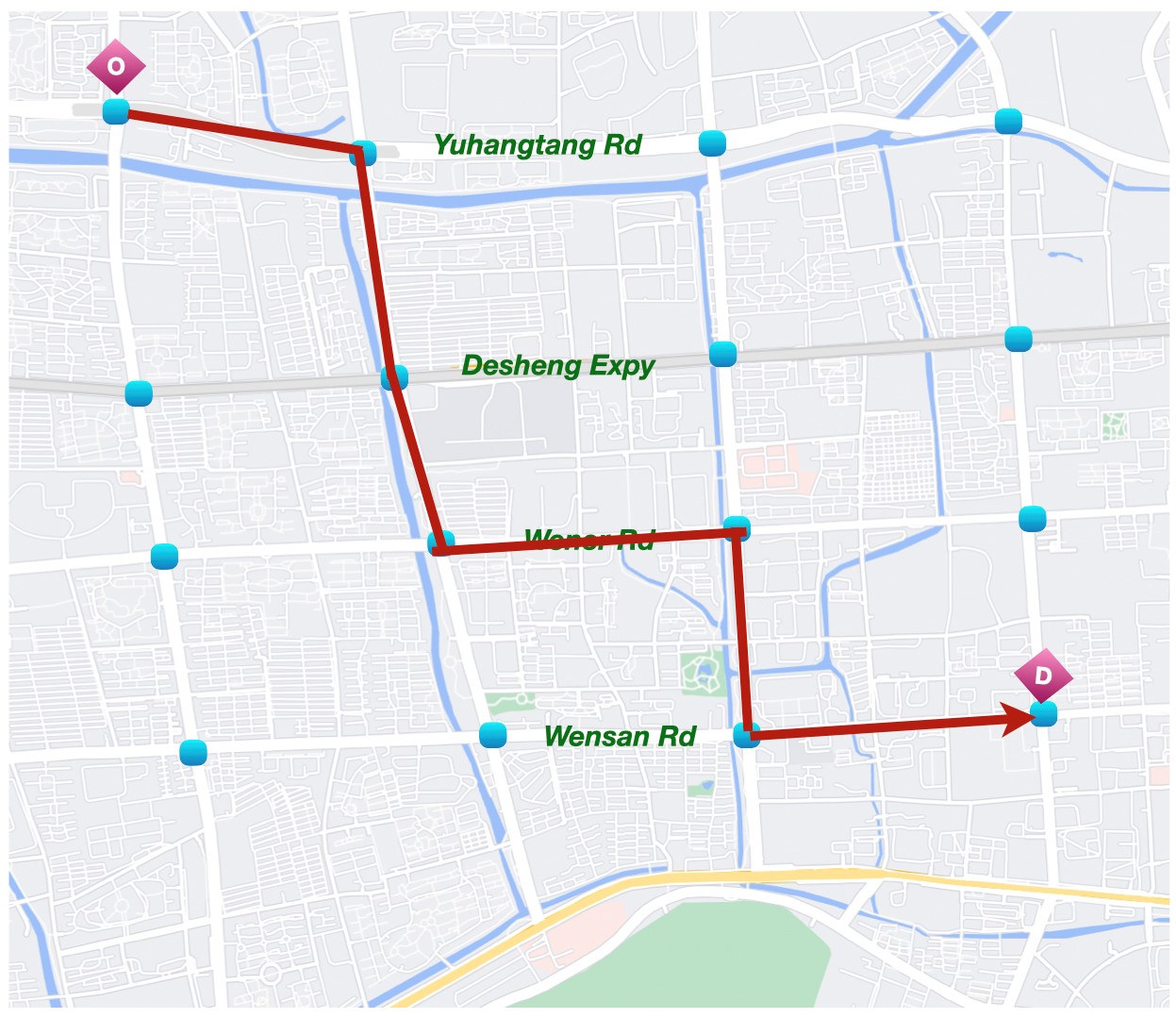}
  \caption{Without primary agents}
  \label{fig_route_choices_ablation_primary}
\end{subfigure}
\caption{EMV's route choice on $\text{Hangzhou}_{4\times 4}$ with replaced primary (a) and secondary (b) agents.}
\label{fig_route_choices_ablation}
\end{figure}

Table \ref{tab_ablation_reward} shows the results of these ablations on the $\text{Hangzhou}_{4\times 4}$ map. We observe that PressLight-style pressure yields a slightly larger $T_{\textrm{EMV}}$ but significantly increases the $T_{\textrm{avg}}$. Without secondary pre-emption agents, $T_{\textrm{EMV}}$ increases by 60\% since almost no ``link reservation" happened. Moreover, without primary pre-emption agents, $T_{\textrm{EMV}}$ increases considerably, which again proves the importance of pre-emption. We can further confirm the importance of agent designs by inspecting the selected routes in the last two ablation studies. 
Fig. \ref{fig_route_choices_ablation_secondary} and \ref{fig_route_choices_ablation_primary} shows routes selected by EMVLights after replacing secondary agents and primary agents with normal agents, respectively. 
Even though the routes are similar as that in Fig.~\ref{fig_emvlight}, much fewer emergency lanes are successfully formed. This failure of utilizing emergency capacity lead to the significant increase in EMV travel time as shown in Table~\ref{tab_ablation_reward}.
As we can see from the routes, EMVs barely take advantage of emergency yielding during their trips.


\subsubsection{Ablation study on policy exchanging}
\begin{figure}[ht]
    \centering
    \includegraphics[width=0.9\linewidth]{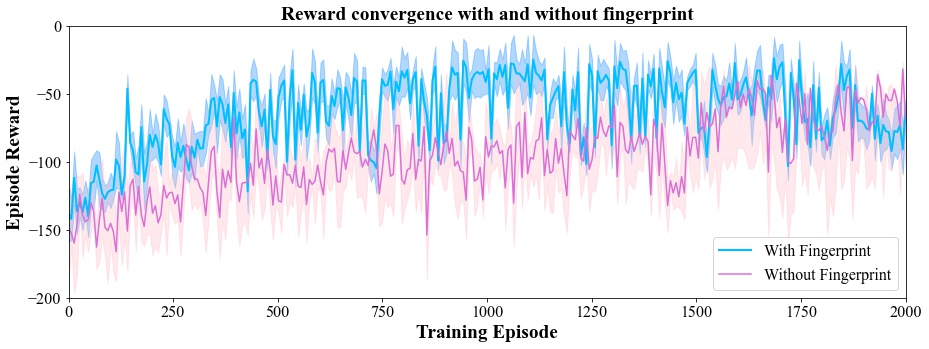}
  \caption{Reward convergence with and without fingerprint. Experiments are conducted on Config 1 synthetic $\text{grid}_{5 \times 5}$.}
  \label{fig_FP_comparison}
\end{figure}

In multi-agent RL, fingerprint has been shown to stabilize training and enable faster convergence.
In order to see how fingerprint affects training in EMVLight, we remove the fingerprint design, i.e., policy and value networks are changed from $\pi_{\theta_i}(a_i^t|s^t_{\mathcal{V}_i}, \pi^{t-1}_{\mathcal{N}_i})$ and  $V_{\phi_i}(\Tilde{s}^t_{\mathcal{V}_i}, \pi^{t-1}_{\mathcal{N}_i})$ to $\pi_{\theta_i}(a_i^t|s^t_{\mathcal{V}_i})$ and  $V_{\phi_i}(\Tilde{s}^t_{\mathcal{V}_i})$, respectively. 
Fig.~\ref{fig_FP_comparison} shows the influence of fingerprint on training. With fingerprint, the reward converges faster and suffers from less fluctuation, confirming the effectiveness of fingerprints, i.e. policy exchanging.


\section{Conclusion}\label{sec_conclusion}
In this paper, we introduced a decentralized reinforcement learning framework, EMVLight, to facilitate the efficient passage of EMVs and reduce traffic congestion at the same time.
Leveraging the multi-agent A2C framework, agents incorporate dynamic routing and cooperatively control traffic signals to reduce EMV travel time and average travel time of non-EMVs.
Our work considers the realistic settings of emergency capacitated road segments and traffic patterns before, during and after EMV passages. Extending Dijkstra's algorithm and embedding into the multi-class RL agent design, EMVLight fundamentally addresses the coupling challenge of EMV's dynamic routing and traffic signal control, filling the research gap on this particular task.
Evaluated on both synthetic and real-world map, EMVLight shortens the EMV travel time and average travel time by up to $42.6\%$ and $23.5\%$ respectively, comparing with existing methods from traditional and learning-based traffic signal control for EMV-related managements. Both quantitative and qualitative assessments on EMVLight, including its EMV navigation as well as pre-clear and restore traffic conditions under emergency state, have concluded that EMVLight is a promising control scheme for such task.

Non-trivial future directions for this study including, but not limited to, the followings. First, the interactions among vehicles, especially under emergency, are extremely complicated. As an effort to close the gap between simulation and reality, we are motivated to extend current ETA estimation model to capture more realistic traffic patterns with EMVs so that agents are more responsive in the field tests. Second, we are looking forward to navigating multiple EMVs simultaneously in the same traffic network. The reward design for pre-emption agents (imagine two or more EMVs appears within one intersection) is worth a technical and ethical discussion. Last but not least, as one of the RL applications in the field of transportation, our method has required a massive number of updated iterations, even in the simulated environments, to achieve the desirable result. How to learn efficiently so that our trail-and-error attempts would not bring catastrophic impacts on real traffic is a critical question for all ITS RL applications.

\section*{CRediT authorship contribution statement}
\textbf{Haoran Su}: Conceptualization, Investigation, Methodology, Visualization, Validation, Writing - original draft, Final approval of the version to be submitted,
\textbf{Yaofeng D. Zhong}: Conceptualization, Investigation, Methodology, Validation, Writing - review \& editing, Final approval of the version to be submitted, 
\textbf{Joseph Y.J. Chow}: Conceptualization, Methodology, Supervision, Validation, Writing - review \& editing, Final approval of the version to be submitted,
\textbf{Dey Biswadip}: Conceptualization, Methodology, Validation, Writing - review \& editing, Final approval of the version to be submitted,
\textbf{Li Jin}: Conceptualization, Methodology, Supervision, Validation, Writing - review \& editing, Final approval of the version to be submitted. 

\section*{Declaration of Competing Interest}
The authors declare that they have no known competing financial interests or personal relationships that could have appeared to influence the work reported in this paper.

\section*{Acknowledgements}
The authors are thankful to Amit Chakraborty for discussion on the conceptualization. 
This work was in part supported by Siemens Corporation, Technology,  
Dwight David Eisenhower Transportation Fellowship, C2SMART University Transportation Center, NSFC Project 62103260, SJTU UM Joint Institute, and J. Wu \& J. Sun Endowment Fund..
These supports are gratefully acknowledged, but imply no endorsement of the findings.
\bibliographystyle{elsarticle-num-names}
\bibliography{_main.bib}
\newpage
\appendix
\appendixpage

\section{Implementation Details}\label{appendix_a}
\textbf{MDP time step.} Although MDP step length can be arbitrarily small enough for optimality, traffic signal phases should be maintained a minimum amount of time so that vehicles and pedestrians can safely cross the intersections. To avoid rapid switching between the phases, we set our MDP time step length to be 5 seconds.


\textbf{Implementation details for Non-emergency-capacitated/Emergency capacitated Synthetic \texorpdfstring{$\text{Grid}_{5 \times 5}$}{Grid}}
\begin{itemize}
    \item dimension of $s^t_{\mathcal{V}_i}$: $5 \times (8+8+4+2)=110$
    \item dimension of $\Tilde{s}^t_{\mathcal{V}_i}$: $5 \times (8+8+4+2)=110$
    \item dimension of $\pi^{t-1}_{\mathcal{N}_i}$: $4 \times 8 = 32$
    \item Policy network $\pi_{\theta_i}(a_i^t|s^t_{\mathcal{V}_i}, \pi^{t-1}_{\mathcal{N}_i})$: 
    \texttt{concat[}$110 \xrightarrow[]{\textrm{FC}} 128$ReLu, $32 \xrightarrow[]{\textrm{FC}} 64$ReLu\texttt{]} $ \xrightarrow[]{} 64$LSTM $ \xrightarrow[]{\textrm{FC}}8$Softmax
    \item Value network $V_{\phi_i}(\Tilde{s}^t_{\mathcal{V}_i}, \pi^{t-1}_{\mathcal{N}_i})$: \texttt{concat[}$110 \xrightarrow[]{\textrm{FC}} 128$ReLu, $32 \xrightarrow[]{\textrm{FC}} 64$ReLu\texttt{]} $ \xrightarrow[]{} 64$LSTM $ \xrightarrow[]{\textrm{FC}}1$Linear
    \item Each link is $200m$. The free flow speed of the EMV is $12m/s$ and the free flow speed for non-EMVs is $6m/s$.
    \item Temporal discount factor $\gamma$ is $0.99$ and spatial discount factor $\alpha$ is $0.90$.
    \item Initial learning rates $\eta_\phi$ and $\eta_\theta$ are both 1e-3 and they decay linearly. Adam optimizer is used.
    \item MDP step length $\Delta t = 5s$ and for secondary pre-emption reward weight $\beta$ is $0.5$.
    \item Regularization coefficient is $0.01$.
\end{itemize}
\textbf{Implementation details for \texorpdfstring{$\text{Manhattan}_{16 \times 3}$}{Manhattan}}
The implementation is similar to the synthetic network implementation, with the following differences:
\begin{itemize}
    \item Initial learning rates $\eta_\phi$ and $\eta_\theta$ are both 5e-4.
    \item Since the avenues and streets are both one-directional, the number of actions of each agent are adjusted accordingly. 
    \item Avenues and streets length are based on real Manhattan block size with each block of $80m \times 274m$.
\end{itemize}

\textbf{Implementation details for \texorpdfstring{$\text{Hangzhou}_{4 \times 4}$}{Manhattan}}
The implementation is similar to the synthetic network implementation, with the following differences:
\begin{itemize}
    \item Initial learning rates $\eta_\phi$ and $\eta_\theta$ are both 5e-4.
    \item Streets length are based on real map of Hangzhou Gudang district.
\end{itemize}

\section{Hyperparameters}\label{appendix_b}
We provide the the choice of hyper-parameters for RL-based methods in Table.\ref{tab_RL_hyperparameters}.
\begin{table}[ht]
\centering
\fontsize{9.0pt}{10.0pt} \selectfont
\begin{tabular}{@{}ccccc@{}}
\toprule
Hyper-parameters & CDRL   & PL   & CL   & EMVLight   \\ \midrule
temporal discount  &   \multicolumn{4}{c}{0.99} \\
batch size      &  32     &  128    & 128     &    128        \\
buffer size     & 1e5      &  1e4    &  1e4    &   1e4         \\
sample size     &  2048     & 1000     & 1000     &  1000          \\
$\eta$ and decay  & 0.5\&0.975 & 0.8\&0.95     &  0.8\&0.95    & 0.8\&0.95           \\ \midrule
optimizer       &  Adam     & RMSprop     &  RMSprop    &   Adam         \\
Learning rate   &  0.00025     &  1e-3    & 1e-3     &   1e-3         \\ \midrule
\# Conv layers   &   1    &  -     &  3    &    -        \\
\# MLP layers    &   1    &   4   &  3    &      1      \\
\# MLP units     &   (168,168)    & (32,32)     &  (32,32)    &   (192,1)         \\
MLP Activation  & \multicolumn{4}{c}{ReLU}         \\
Initialization  & \multicolumn{4}{c}{Random Normal} \\
\midrule
step length $\Delta T$      &   \multicolumn{4}{c}{ 5 seconds}      \\
\bottomrule
\end{tabular}
\caption{Hyper-parameters selected for RL-based methods.}
\label{tab_RL_hyperparameters}
\end{table}






\end{document}